\documentclass[runningheads]{llncs}

\usepackage[mobile]{eccv}

\usepackage{times}
\usepackage{epsfig}
\usepackage{graphicx}
\usepackage{amsmath}
\usepackage{amssymb}
\usepackage{booktabs}
\usepackage{mwe}
\usepackage{pdfpages}
\usepackage{multirow}
\usepackage{rotating}
\usepackage{adjustbox}
\usepackage{tabularx}
\usepackage{array}
\usepackage{dsfont}

\usepackage{eccvabbrv}

\usepackage{graphbox}
\usepackage{graphicx}
\usepackage{booktabs}
\newcommand\blfootnote[1]{%
  \begingroup
  \renewcommand\thefootnote{}\footnote{#1}%
  \addtocounter{footnote}{-1}%
  \endgroup
}
\usepackage[accsupp]{axessibility}  %

\usepackage[pagebackref,breaklinks,colorlinks,citecolor=eccvblue]{hyperref}

\usepackage{orcidlink}

\newcommand{\ourM}{SegVLAD}

\begin{document}

\title{Revisit Anything: Visual Place Recognition via Image Segment Retrieval} 

\titlerunning{Revisit Anything}

\author{Kartik Garg*  \inst{1}\orcidlink{0000-0002-8585-3939} \and
Sai Shubodh Puligilla* \inst{2}\orcidlink{0009-0000-7643-4479} \and 
Shishir Kolathaya \inst{1} \orcidlink{0000-0001-8689-2318}\and \\
Madhava Krishna \inst{2} \orcidlink{0000-0001-7846-7901}\and
Sourav Garg \inst{3} \orcidlink{0000-0001-6068-3307}}
\authorrunning{K.~Garg et al.}

\institute
{Indian Institute of Science (IISc), Bengaluru, India \and
International Institute of Information Technology, Hyderabad, India \and
University of Adelaide, Australia\\
  \blfootnote{* Equal contribution}
}

\maketitle

\begin{abstract}
  Accurately recognizing a revisited place is crucial for embodied agents to localize and navigate. This requires visual representations to be distinct, despite strong variations in camera viewpoint and scene appearance. Existing visual place recognition pipelines encode the \textit{whole} image and search for matches. 
  This poses a fundamental challenge in matching two images of the same place captured from different camera viewpoints: \textit{the similarity of what overlaps can be dominated by the dissimilarity of what does not overlap}. 
  We address this by encoding and searching for \textit{image segments} instead of the whole images. We propose to use open-set image segmentation to decompose an image into `meaningful' entities (i.e., things and stuff). This enables us to create a novel image representation as a collection of multiple overlapping subgraphs connecting a segment with its neighboring segments, dubbed SuperSegment. Furthermore, to efficiently encode these SuperSegments into compact vector representations, we propose a novel factorized representation of feature aggregation. We show that retrieving these partial representations leads to significantly higher recognition recall than the typical whole image based retrieval. Our segments-based approach, dubbed \ourM{}, sets a new state-of-the-art in place recognition on a diverse selection of benchmark datasets, while being applicable to \textit{both} generic and task-specialized image encoders. Finally, we demonstrate the potential of our method to ``revisit anything'' by evaluating our method on an object instance retrieval task, which bridges the two disparate areas of research: visual place recognition and object-goal navigation, through their common aim of recognizing goal objects specific to a place. 
  \\Source code: \url{https://github.com/AnyLoc/Revisit-Anything}.

\end{abstract}

\section{Introduction}
Visual Place Recognition (VPR) is an important capability for embodied agents to localize and navigate autonomously. A predominant solution for VPR is to encode an image into a global vector and retrieve similar vectors as coarse localization hypotheses~\cite{schubert2023visual,garg2021where,tsintotas2022revisiting,masone2021survey}. 
Thus, for almost a decade, researchers have focused on learning/finetuning image encoders so that global descriptors are induced with invariance to appearance~\cite{arandjelovic2016netvlad,warburg2020mapillary,revaud2019learning}, viewpoint~\cite{berton2023eigenplaces,arandjelovic2016netvlad}, and clutter~\cite{ibrahimi2021inside}. On the other hand, there is a vast literature on local descriptors (point/pixel-level), mainly relevant for geometric reranking in hierarchical VPR~\cite{cummins2011appearance,sarlin2018acoarse,hausler2021patchnetvlad,cao2020unifying,wang2022transvpr}. In the middle of local and global descriptors exists a variety of methods that use regions/patches~\cite{hausler2021patchnetvlad,arandjelovic2013all}, lines/planes~\cite{cupec2015place}, objects (things/stuff)~\cite{mirjalili2023fm, kassab2023clip,sunderhauf2015place,garg2018lost}, and segments~\cite{hu2020dasgil,paolicelli2022learning,keetha2021hierarchical} to represent images. However, these methods are still only aimed at either improving global descriptors based coarse retrieval or local feature matching based reranking. In this work, in contrast to conventional retrieval-based VPR, we explore an alternative: \textbf{retrieval via encoding segments instead of the whole image}. This is particularly enabled by recent advances in open-set image segmentation~\cite{kirillov2023segment} which can meaningfully deconstruct a place into `things' (and/or `stuff')~\cite{caesar2018coco}. Thus, we reformulate the VPR problem of revisiting places as that to \textit{revisiting things} by enabling recognition of these specific things within the context of their place. While such a segment-level place recognition approach provides a direct link to higher-level semantic tasks, such as object-goal navigation~\cite{chang2023goat,garg2024robohop,gu2023conceptgraphs,manglani2023real}, it also addresses a fundamental issue in matching partially-overlapping images from across significant viewpoint change. Segments-based partial image representation avoids the mismatches caused by the whole-image representation when \textit{the similarity of what overlaps is dominated by the dissimilarity of what does not overlap}. Our novel segments-based VPR method, dubbed \textit{\ourM{}} (Segment based Vector of Locally Aggregated Descriptors), is illustrated in Figure~\ref{fig:pipeline}, which makes the following novel contributions:
\begin{enumerate}
    \item an image representation as a collection of multiple overlapping subgraphs of segments, dubbed \textit{SuperSegments}, which enables accurate recognition across partially-overlapping images;
    \item a factorized representation of feature aggregation to effectively accommodate both segment-level information as well as segment neighborhood information; and
    \item a similarity-weighted ranking method to convert segment-level retrieval into image-level retrieval.
\end{enumerate}

Using a diverse set of data sources, we demonstrate that our proposed segments-based retrieval enables place recognition under wide viewpoint variations, where global descriptor based retrieval suffers. \ourM{} achieves a new state-of-the-art on multiple challenging datasets. We also introduce an evaluation of our method on an instance-level object retrieval task -- a novel capability of our pipeline unlike conventional VPR methods. We conduct several ablations and parameter studies to justify the design choices and emphasize the effectiveness of our method as an open-set segments-based coarse retriever.

\section{Related Works}
Image retrieval-based Visual Place Recognition (VPR) is a well-established area of research in visual localization~\cite{garg2021where,masone2021survey,schubert2023visual,tsintotas2022revisiting}. It is important both during mapping for loop closures~\cite{tsintotas2022revisiting} as well as for relocalization~\cite{sattler2018benchmarking,pion2020benchmarking}. The underlying task in both the scenarios remains the same: how to recognize a previously seen place. The state-of-the-art methods in VPR use a global descriptor-based approach which converts an image into a compact vector to enable fast retrieval~\cite{arandjelovic2016netvlad,revaud2019learning,berton2022rethinking,ali2023mixvpr,keetha2023anyloc,izquierdo2023optimal}. The top retrieved hypotheses are often then re-ranked through compute-intensive local feature matching using geometric information~\cite{cummins2011appearance,hausler2021patchnetvlad,wang2022transvpr,keetha2021hierarchical,sarlin2018acoarse,cao2020unifying}. In contrast to previous approaches, we aim to explore image segment level descriptors in this work. This representation falls between point-based local descriptors and the whole-image based global descriptors. Our approach can be considered as `semi-global', with the proposed segment (and SuperSegment) based descriptors being a `spatially-reduced' form of whole-image global representation. This is motivated by our hypothesis that to deal with viewpoint variations in VPR with partially-overlapping images, we need a way to partially represent and match them.    

\subsection{Whole Image Encoders}
Earlier works in whole-image representation used methods like Gist~\cite{oliva2005gist}, BoW (Bag of Word)~\cite{sivic2003video}, and VLAD (Vector of Locally Aggregated Descriptors)~\cite{jegou2010aggregating}, often defined using hand-crafted features such as SIFT~\cite{lowe2004distinctive}. In recent years, deep learning based methods have demonstrated remarkable performance, with initial successful methods like NetVLAD~\cite{arandjelovic2016netvlad} now rapidly outperformed by better alternatives such as CosPlace~\cite{berton2022rethinking}, MixVPR~\cite{ali2023mixvpr}, EigenPlaces~\cite{berton2023eigenplaces}, TransVPR~\cite{wang2022transvpr}, and more recently AnyLoc~\cite{keetha2023anyloc}, SALAD~\cite{izquierdo2023optimal} and VLAD-BuFF~\cite{khaliq2024vladbuff}. All these learning-based methods improve different aspects of representation learning: training datasets~\cite{berton2022rethinking,warburg2020mapillary,ali2022gsv}, objective/loss functions~\cite{berton2023eigenplaces,ali2023mixvpr}, aggregation methods~\cite{izquierdo2023optimal,radenovic2018fine,revaud2019learning,wang2022transvpr}, and generalization~\cite{keetha2023anyloc}. Our approach complements these existing methods as we mainly focus on the use of segment-based information, where the segments can be described by any of the image encoders from the aforementioned techniques. In particular, we demonstrate that \textit{both} -- an off-the-shelf encoder, e.g., DINOv2-AnyLoc~\cite{oquab2023dinov2,keetha2023anyloc} or that finetuned specifically for the VPR task, e.g., DINOv2-NetVLAD~\cite{khaliq2024vladbuff} -- can be used in conjunction with our segment-based approach to further elevate place recognition capability.  

\subsection{Region/Patch Based Methods}
There exist several methods that employ region or patch level information to enhance representational power~\cite{yu2019spatial, le2020city, chen2017only, khaliq2018holistic, chen2016attention, puligilla2020topological, xin2019localizing}. However, most of these methods only use this additional information to generate a single (or concatenated) compact vector representation of an image. Other methods such as Patch-NetVLAD~\cite{hausler2021patchnetvlad} create multiple features per image but their primary purpose is to perform local matching based reranking. In contrast to these methods, we aim to use multiple segment descriptors per image to directly retrieve from a database of segments, \textit{without} using any geometric information or reranking. The motivation behind this stems from the very nature of hierarchical VPR pipelines: reranking recall is upper bounded by the coarse retriever. A better coarse retriever can improve reranking performance without needing to rerank from a longer list of top hypotheses. MultiVLAD~\cite{arandjelovic2013all} is similar to our method in the spirit of retrieving multiple features per query image. However, like aforementioned methods, MultiVLAD defines regions arbitrarily, whereas we use image segments obtained from Segment Anything Model (SAM)~\cite{kirillov2023segment} which are semantically meaningful.

\subsection{Segments-Enhanced Methods}
There exist several methods that use semantic segmentation information to improve VPR, as also surveyed in~\cite{garg2020semantics}. These methods vary in terms of type of segmentation used and the specific ways in which it is integrated in the VPR pipeline, e.g., planes~\cite{cupec2015place}, objects~\cite{cheng2008object,mirjalili2023fm, kassab2023clip}, landmarks~\cite{sunderhauf2015place}, outdoor semantics~\cite{garg2018lost,gawel2018x,naseer2017semantics,mousavian2016semantic}, utility/confusion based~\cite{knopp2010avoiding,keetha2021hierarchical}, domain adaptation~\cite{hu2020dasgil} and even learning to segment for VPR~\cite{paolicelli2022learning}. However, neither these methods aim to perform segment-level retrieval nor do they use open-set segmentation. 
We also review two concurrent works:
MESA~\cite{zhang2024mesa} and Region-Revisited~\cite{shlapentokh2024region}. Similar to our method, they both use SAM to segment images but for different specific tasks. MESA~\cite{zhang2024mesa} proposes a graph-based local feature/area matching method to obtain point correspondences. Our method complements this, as we perform coarse retrieval for VPR, which could potentially use MESA for reranking. Regions Revisted~\cite{shlapentokh2024region} delves into the advantages of using SAM masks in conjunction with SLIC~\cite{khorasgani2022slic} to improve semantic segmentation, activity recognition and object \textit{category} retrieval. In contrast, we aim to improve \textit{instance-level} recognition by recognizing specific things belonging to specific places that a robot encounters during a revisit. Similar to our work,~\cite{garg2024robohop} creates an image sequence-based topological graph of segments where its segment neighbourhood aggregation is based on average pooling, similar to~\cite{shlapentokh2024region}. In Section~\ref{sec:ablate}, we show that such segment average pooling deteriorates recognition performance for the VPR task.

\subsection{Open-set VPR}
Researchers have recently started to shift their focus to open-set, generally-applicable techniques, including that for VPR~\cite{keetha2023anyloc,mirjalili2023fm, kassab2023clip, shubodh2024lip}. FM-Loc~\cite{mirjalili2023fm} uses GPT~\cite{brown2020language} to recognize object and place labels, whereas~\cite{kassab2023clip} uses CLIP~\cite{radford2021learning} for open-set place recognition. LIP-Loc~\cite{shubodh2024lip} proposes pretraining for cross-modal VPR, but is limited in its zero-shot capabilities. AnyLoc~\cite{keetha2023anyloc} proposes to use DINOv2 with domain-level vocabularies and hard-assignment based VLAD. It achieves state-of-the-art performance particularly on non-streetview datasets, where current VPR-trained methods tend to fail. In this work, we propose a generally-applicable approach which is built on top of models like SAM~\cite{kirillov2023segment} and DINOv2~\cite{oquab2023dinov2}, and works with both VPR-agnostic~\cite{keetha2023anyloc} and VPR-specific~\cite{khaliq2024vladbuff} backbone models. We particularly aim for a new paradigm in retrieval based VPR, where we move away from the conventional whole image global descriptors to segments based descriptors and retrieval, which achieves a new state-of-the-art on diverse domains under wide viewpoint variations. 

\begin{figure}[t]
    \centering
    \includegraphics[width=\textwidth,trim={4cm 7cm 5cm 1cm},clip]{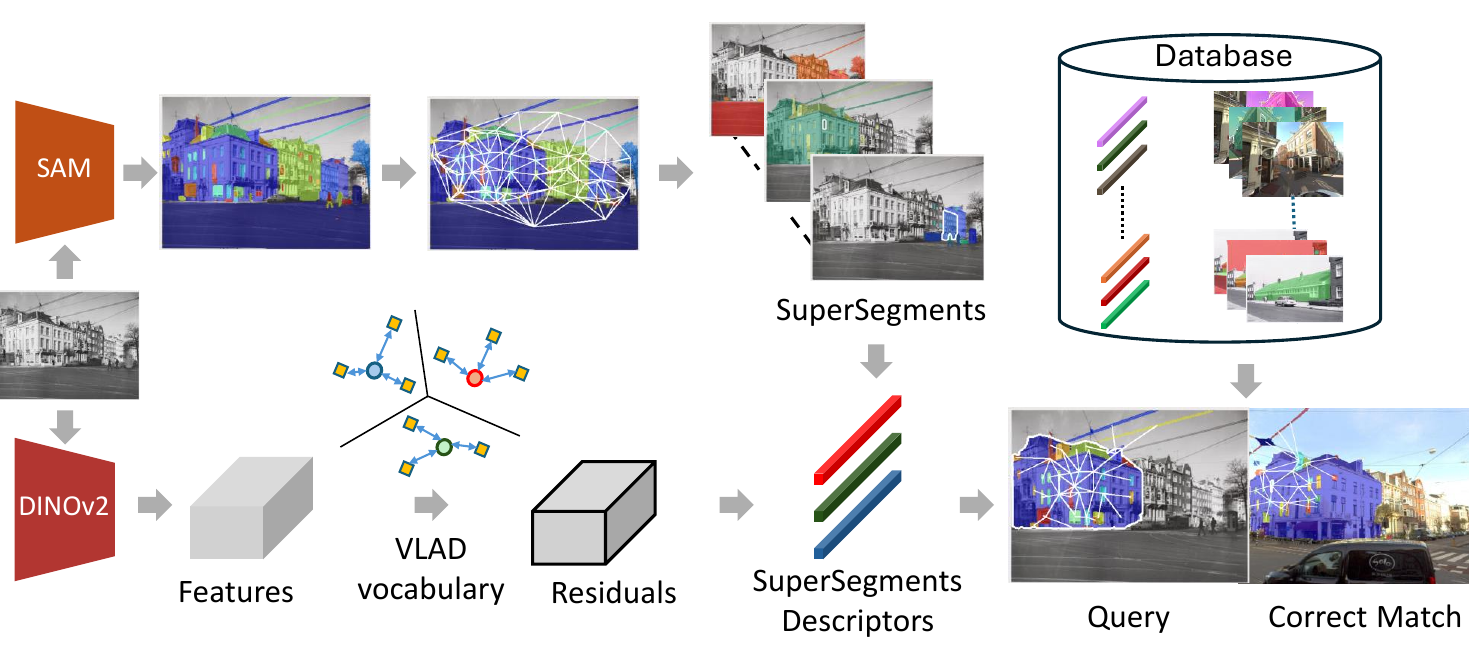}
    \caption{Overview of our segment-retrieval based VPR pipeline, dubbed \ourM{}: We use open-set segmentor (SAM) to extract segment masks, which are converted into \textbf{SuperSegments} using the neighbouring image segments. Using pixel-level DINOv2 descriptors with VLAD-based aggregation, we obtain SuperSegment descriptors, which are matched against a flat index of SuperSegment descriptors obtained from all the images of the entire reference database.}
    \label{fig:pipeline}
\end{figure}

\section{Proposed Approach} \label{sec:approach}
Despite recent advances in place recognition, viewpoint variations continue to be an open challenge for an embodied agent to recognize the same specific things in its environment. Current methods in visual place recognition tackle this problem by converting an image \textit{as a whole} into a global descriptor, which does not explicitly deal with the problem of partial visual overlap caused by viewpoint variations. We propose an alternative solution by representing images \textit{partially} with the help of image segments. In the following subsections, we describe our representation and retrieval method, which deviates from the conventional VPR techniques but creates a new capability in terms of recognizing objects/things that constitute a place. 
    
\subsection{Problem Formulation}
We represent an image as a set of segment descriptors instead of a single global descriptor. For an image $I$, we obtain binary image segment masks $M \in \{0,1\}^{S\times N}$ and dense pixel-level descriptors $f_{p} \in \mathbb{R}^D$, where $S$ represents the number of segments per image, $D$ is the descriptor dimension, and $p \in [1,N]$ represents spatial elements across the width ($W$) and height ($H$) of the image encoder's output, flattened into $N=W\times H$ for convenience. Figure~\ref{fig:pipeline} shows an illustration of our proposed pipeline, as explained in the following subsections.

\subsection{Super Segments}\label{sec:seg_agg}

\begin{figure}[t]
    \centering
         \resizebox{\columnwidth}{!}{%
    \begin{tabular}{cccc}
         \includegraphics[width=.5\linewidth]{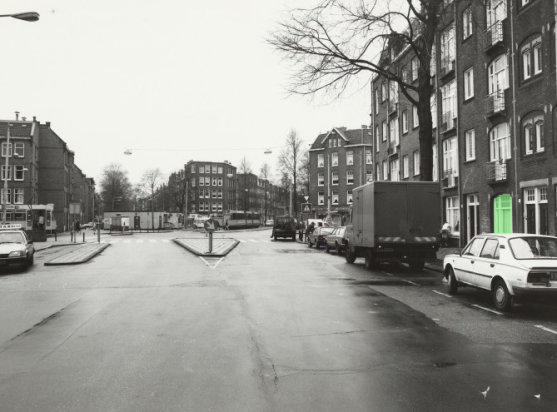}& \includegraphics[width=.5\linewidth]{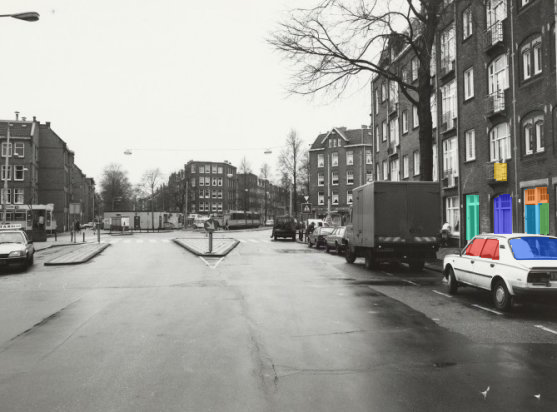} & \includegraphics[width=.5\linewidth]{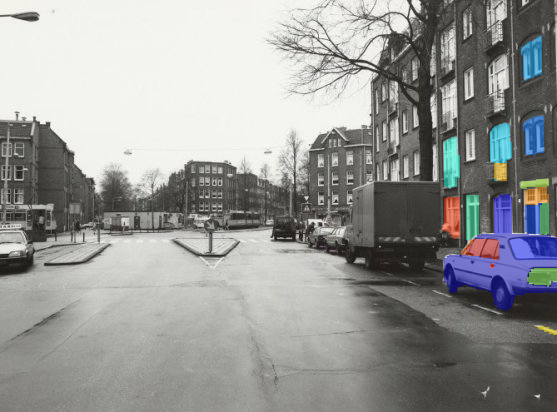} & \includegraphics[width=.5\linewidth]{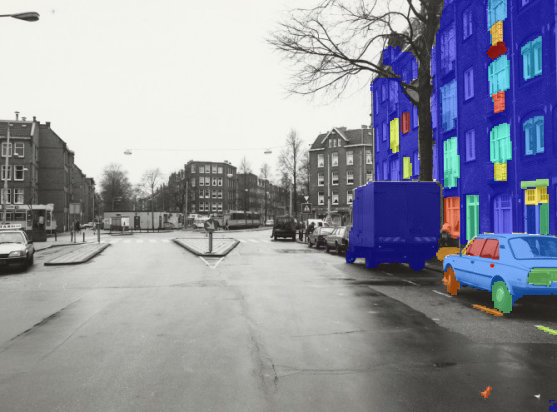}\\
    \end{tabular}%
    }
    \caption{Neighborhood expansion (Eq.~\ref{eq:ss}) of a window in the leftmost image to the whole building in the rightmost image, progressing from no neighborhood aggregation to a third-order aggregation. This neighborhood expansion is in stark contrast with a typical regular grid- or patch-based approach which may not capture semantically-meaningful SuperSegments.}
    \label{fig:nbragg}
\end{figure}

\begin{figure}[t]
    \centering
         \resizebox{\columnwidth}{!}{%
    \begin{tabular}{cccc}
         \includegraphics[width=.5\linewidth]{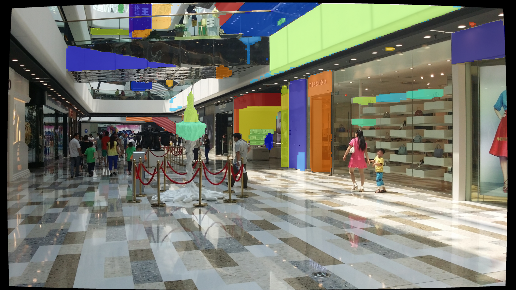}& \includegraphics[width=.5\linewidth]{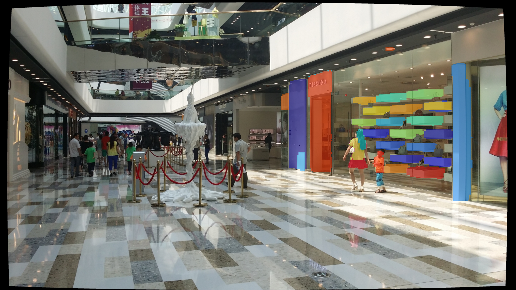} & \includegraphics[width=.5\linewidth]{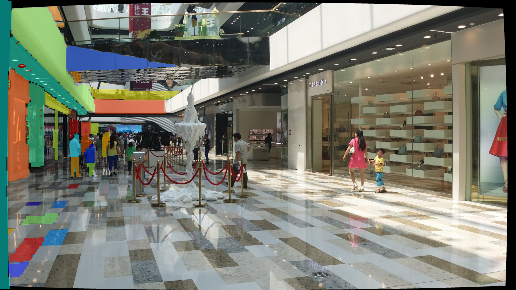} & \includegraphics[width=.5\linewidth]{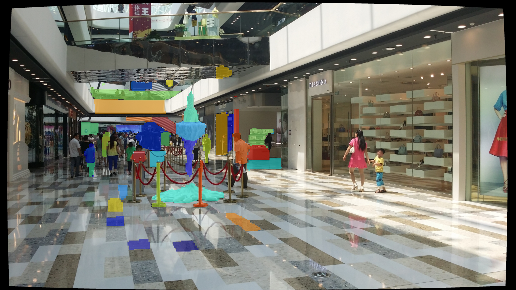}\\
    \end{tabular}%
    }
    \caption{Illustration of four SuperSegments obtained from the same image. All four of these spatially overlap with each other, which is different from coarse segmentation methods that do not typically allow overlap across segments.}
    \label{fig:overlap}
\end{figure}

Humans are remarkable at visual recognition, where existing studies suggest that we often leverage spatial associations among objects in an environment to represent it internally~\cite{bar2004visual, intraub1997representation}. This enables us to distinguish between two different scenes through the surrounding context of the objects of interest. In this work, we imbibe this context through the spatial neighbourhood of the image segments.
For each image, we construct a graph of segments through their pixel centers using Delaunay Triangulation. This provides us with a binary adjacency matrix $A \in \{0,1\}^{S \times S}$ to define the neighborhood for individual segments. We use this adjacency information to expand the context of individual segments to generate new \textbf{SuperSegment} masks ($\mathcal{M}$) as below:
\begin{equation}
    \mathcal{M}_{S\times N} =  {}^{\mathds{1}}(A_{S\times S}^o \cdot M_{S\times N})\
    \label{eq:ss}
\end{equation}
where $o\geq0$ refers to the order for expanding the neighborhood by multiplying the adjacency matrix by itself as $A^{o+1}=A^o \cdot A$. This is matrix-multiplied with the original segmentation masks $M$ to expand the neighborhood \textit{at pixel level}. $\mathcal{M}$ is obtained after element-wise binarization (denoted with ${}^{\mathds{1}}()$) so that all pixels in the SuperSegment mask may only contribute once to the subsequent feature aggregation. In Figure~\ref{fig:nbragg}, we illustrate the extent of image area covered with different orders of mask expansion. Unlike, a patch or regular grid-based approach, the expanded mask of the window in the leftmost image covers a meaningful entity (building) in the rightmost image.
Our approach to creating SuperSegments differs from \textit{coarse} segmentation methods or superpixels in terms of the `self-overlap'. By expanding neighborhood of each individual segment, we obtain several \textit{partially overlapping} SuperSegments. A coarse segmentor will need to make assumptions about the right sub-segments to be coalesced so that it can enable accurate recognition from a different viewpoint, which could otherwise lead to the same limitation as that of the whole-image descriptors. Figure~\ref{fig:overlap} presents examples of multiple overlapping SuperSegments from the same image.

\subsection{SuperSegment Descriptors} \label{sec:seg_feat}
In this section, we describe our feature aggregation method to obtain SuperSegment descriptors. Recent state-of-the-art method AnyLoc~\cite{keetha2023anyloc} demonstrated that using off-the-shelf powerful image encoders such as DINOv2 with hard assignment based VLAD aggregation achieves superior recognition performance. However, AnyLoc does not use segmentation information and only operates at the whole-image level. More recently,~\cite{shlapentokh2024region} showed that average pooling works well for segment-level descriptors, but it didn't consider segment neighborhood information. In this work, we propose a unified formulation for feature aggregation that can easily switch across segments, segment neighborhood and the whole image as well as different aggregation types (see supplementary for details). This simply extends Equation~\ref{eq:ss} as below:
\begin{equation}
    F_{S\times D} = {}^{\mathds{1}}(A_{S\times S}^o \cdot M_{S\times N})\ \cdot T_{N\times D}
    \label{eq:agg}
\end{equation}
where $T$ represents the features to be aggregated. By replacing $A$ and $M$ with ones matrices, one can obtain the whole-image global descriptor for $S=1$. For methods like Global Average Pooling (GAP), $T$ can be directly used as the output of the image encoder. In our work, we use Hard-VLAD, for which $T$ is the residual feature matrix \textit{per cluster} and is obtained as below with respect to each of the cluster centers $c_k$:
\begin{equation}
    T^k_{N_k\times D} = \{\alpha_k(f_{p})(f_p-c_k) \; | \; \alpha_k(f_p) = 1\}; \quad N_k = \sum_p \alpha_k(f_p)
    \label{eq:vlad}
\end{equation}
where $\alpha_k(f_p) \in \{0,1\}$ is 1 if $f_p$ belongs to $c_k$, and . The cluster centers (vocabulary) can be constructed using the map or the domain~\cite{keetha2023anyloc}. The SuperSegment VLAD descriptors obtained from Eq.~\ref{eq:agg} for each cluster center $k$ are l2-normalized per cluster (i.e., intra-normalization), concatenated across clusters and then finally l2-normalized, following existing works~\cite{arandjelovic2016netvlad,keetha2023anyloc}.

\subsection{Image Retrieval via Segments}\label{sec:app_seg}
Existing global descriptor based VPR techniques produce a single vector per image to search against a database of reference image vectors. In our method, we obtain multiple SuperSegment descriptors per image. We perform retrieval at segment-level, that is, we search for the top matches for each query segment against a flat index of all segments from all the images of the reference database/map. To evaluate in the form of image retrieval-based VPR, we convert the top retrieved segment indices across all segments of a query image into top reference image indices. This is achieved through a weighted frequency measure (i.e., weighted bin/word counting). We first map the top $K'$ retrieved segment indices for each of the query segments $s \in [1,S]$ to their respective reference image indices, denoted with $r$. Then, for each of the unique retrieved image indices $r_j$, we accumulate its segment similarity $\theta$ and then use the cumulative similarity score $\hat\theta$ to rank the image indices to obtain the top image match $r_j^*$:
\begin{equation}
r_j^* = \underset{r_j}{\text{argmax}}\; \hat\theta(r_j); \quad
\hat\theta(r_j) = \sum_{s=1}^S \sum_{k=1}^{K'} \theta_{sk} \cdot \mathds{1}_{\{r_{sk} = r_j\}}
\label{eq:simWtRank}
\end{equation}
In Section~\ref{sec:seg_ret}, we compare our similarity-weighted ranking with other alternatives based on frequency or similarity alone.

\section{Experimental Setup}
\paragraph{\textbf{Datasets}:}
VPR datasets are in abundance, as can be found in several benchmarks including VPR-Bench~\cite{zaffar2021vpr}, Deep Visual GeoLocalization Benchmark~\cite{berton2022deep}, and AnyLoc~\cite{keetha2023anyloc}. In this work, we used a variety of datasets covering both outdoor and indoor environments. Outdoor datasets include Pitts30k~\cite{Pittsburgh-30k_dataset}, AmsterTime~\cite{AmsterTime_dataset_Yildiz_2022}, Mapillary Street Level Sequences (MSLS)~\cite{warburg2020mapillary}, SF-XL~\cite{berton2022rethinking}, VPAir~\cite{vpair_dataset_schleiss2022}, Revisted Oxford5K and Revisited Paris6k~\cite{radenovic2018revisiting} . Indoor datasets include Baidu Mall~\cite{baidu_dataset_Sun2017ADF}, 17Places~\cite{zhou2017places} and InsideOut~\cite{ibrahimi2021inside}. Additional datasets-related details are provided in the supplementary.

\paragraph{\textbf{Evaluation and Benchmarking}:}
We evaluate our method as an image retrieval task using Recall@K metric, where top K$'$ ($=50$) retrieved segments per query segment are used to obtain top K ($=5$) images (see Eq.~\ref{eq:simWtRank}). We compare against the most recent and high-performing VPR baseline methods. This includes CosPlace~\cite{berton2022rethinking}, MixVPR~\cite{ali2023mixvpr} and EigenPlaces~\cite{berton2023eigenplaces}, which are trained on large-scale urban datasets for VPR tasks. We further include two very recent state-of-the-art methods that use DINOv2 as the backbone. These include AnyLoc~\cite{keetha2023anyloc} which uses an \textit{off-the-shelf} DINOv2 model and SALAD~\cite{izquierdo2023optimal} which uses a \textit{finetuned} DINOv2 backbone.  
Given the dichotomy between general-purpose VPR benchmarking of AnyLoc and the typical outdoor-focused benchmarking~\cite{berton2022deep}, we evaluated our method using two different backbones. \textit{a)} \textbf{\ourM{}-PreT:}  we use the same backbone and aggregation as AnyLoc, i.e., off-the-shelf \textit{pretrained} DINOv2 (ViT-G) backbone with hard VLAD assignment, but the key difference is in the use of SuperSegments for our method as opposed to whole-image description of AnyLoc. \textit{b)} \textbf{\ourM{}-FineT:} as our default aggregation method is VLAD, we use a \textit{finetuned} DINOv2 (ViT-B) backbone which is similar to SALAD but we replace its aggregation layer with the original NetVLAD aggregation~\cite{arandjelovic2016netvlad} using 64 clusters, as described in~\cite{khaliq2024vladbuff}. We use this finetuned backbone with hard VLAD based assignment, similar to AnyLoc. 
For both these models, we reduce the descriptor dimensions of the VLAD descriptor to $1024$ using PCA, as commonly done in previous works~\cite{arandjelovic2016netvlad,keetha2023anyloc}. We train PCA transform in a map-specific manner using the database images of the dataset. Following AnyLoc~\cite{keetha2023anyloc}, we report results using two different sources of VLAD vocabulary: map-specific \textit{(M)} and domain-specific \textit{(D)}.

\section{Results}

We first present benchmark comparison of our method against state-of-the-art VPR methods.
This is followed by detailed analysis of our proposed aggregation technique. Lastly, we demonstrate results on a downstream task of Object-of-Interest (OOI) retrieval, showcasing the versatility of our method.

\subsection{State-of-the-art comparisons}

\begin{table*}[t]
    
\caption{Recall@1/5 benchmark comparison on outdoor street-view datasets.}

\centering
\resizebox{0.9\linewidth}{!}{%
    \setlength{\tabcolsep}{3\tabcolsep}
    \begin{tabular}{lccccccccc}
        \toprule
        \textbf{Method} & \textbf{Pitts-} & \textbf{MSLS} & \textbf{MSLS} & \textbf{SF-XL} & \textbf{RO5k} & \textbf{RO5k} & \textbf{RP6k} & \textbf{RP6k} \\
        & \textbf{30K} & \textbf{SF} & \textbf{CPH} & \textbf{Val} & \textbf{Med} & \textbf{Hard} & \textbf{Med} & \textbf{Hard} \\
        \midrule
        CosPlace & 90.4/95.7 & \textbf{93.4}/\textbf{97.5} & 84.9/92.0 & 94.6/97.6 & 85.7/87.1 & 27.1/45.7 & 94.3/95.7 & 7.1/15.7 \\
        MixVPR  & 91.5/95.5 & 91.3/95.9 & 87.1/92.4 & 87.8/93.8 & 68.6/80.0 & 32.9/54.3 & 94.3/100 & 10.0/32.9 \\
        EigenPlaces & 92.6/\underline{96.7} & \underline{92.6}/\underline{97.1} & 87.1/92.8 & \textbf{96.4/98.2} & 85.7/88.6 & 42.8/57.1 & 95.7/98.6 & 4.3/11.4 \\
        AnyLoc  & 87.7/94.7 & 83.4/94.6 & 79.9/89.1 & 84.4/91.9 & {88.6/92.9} & 40.0/58.6 & \underline{97.1}/100 & {11.4}/44.3 \\
        SALAD  & 92.6/96.5 & 91.7/\underline{97.1} & \textbf{92.3}/96.1 & 93.6/97.3 & 82.9/90.0 & 37.1/54.3 & 95.7/98.6 & \underline{14.3}/\textbf{58.6} \\
        \midrule
\textbf{\ourM{}-PreT} (D) & 86.7/94.2 & 88.4/94.2 & 81.7/90.7 & 90.9/96.4 & \underline{90.0}/\textbf{97.1} & \underline{47.1}/\underline{72.9} & \textbf{98.5}/100 & \textbf{18.6}/\underline{55.7} \\
\textbf{\ourM{}-PreT} (M) & 83.9/93.2 & 81.0/92.1 & 76.1/89.2 & 90.2/95.9 & \textbf{92.9}/\underline{95.7} & \textbf{61.4}/\textbf{81.4} & \underline{97.1}/100 & {11.4}/40.0 \\
\textbf{\ourM{}-FineT} (D) & \underline{92.9}/\textbf{96.8} & \textbf{93.4}/\underline{97.1} & \underline{91.8}/\underline{96.4} & 94.2/\underline{97.9} & 82.9/92.9 & 40.0/60.0 & \underline{97.1}/100 & 5.7/52.9 \\
\textbf{\ourM{}-FineT} (M) & \textbf{93.1}/\textbf{96.8} & 92.2/\underline{97.1} & 91.6/\textbf{96.6} & \underline{95.6}/\textbf{98.2} & 84.3/91.4 & 44.3/61.4 & 95.7/100 & 10.0/57.1 \\

        \bottomrule
    \end{tabular}%
}
\label{tab:result_table}

\end{table*}

\begin{table*}[t]

\caption{Recall@1/5 benchmark comparison on `out-of-distribution' datasets.}

\centering
\resizebox{0.65\linewidth}{!}{%
    \setlength{\tabcolsep}{2\tabcolsep}
    \begin{tabular}{lccccc}
        \toprule
        \textbf{Method} & \textbf{Baidu} & \textbf{AmsterTime} & \textbf{InsideOut} & \textbf{17Places} & \textbf{VPAir} \\
        \midrule
        CosPlace & 41.6/55.0 & 47.7/69.8 & 0.2/2.0 & 81.3/88.2 & 4.6/13.7 \\
        MixVPR  & 64.4/80.3 & 40.2/59.1 & 0.0/1.8 & 85.2/90.1 & 6.8/16.1 \\
        EigenPlaces & 56.5/72.8 & 48.9/69.5 & 0.4/1.4 & 83.0/90.1 & 6.5/17.9 \\
        AnyLoc  &{75.2}/87.6 & 50.3/73.0 & 2.4/8.0 & \textbf{95.3}/97.3 & {66.7/79.2} \\
        SALAD  &74.8/86.5 & 55.4/75.6 & 0.6/1.8 & 82.5/88.2 & 25.8/38.7 \\
        \midrule

\textbf{\ourM{}-PreT} (D) & \underline{78.5}/\underline{93.8} & 48.3/72.4 & 4.2/9.4 & \textbf{95.3}/\underline{98.0} & \textbf{69.8}/\textbf{83.7} \\
\textbf{\ourM{}-PreT} (M) & \textbf{80.4}/\textbf{94.0} & 54.3/76.0 & 3.2/10.4 & \textbf{95.3}/\underline{98.0} & \underline{67.2}/\underline{82.8} \\
\textbf{\ourM{}-FineT} (D) & 69.9/89.5 & \underline{56.7}/\underline{76.8} & \underline{7.0}/\underline{14.0} & \textbf{95.3/}97.8 & 33.9/52.4 \\
\textbf{\ourM{}-FineT} (M) & 69.7/90.3 & \textbf{60.2}/\textbf{78.2} & \textbf{7.2}/\textbf{17.2} & \textbf{95.3}/\textbf{98.3} & 34.2/52.1 \\

        \bottomrule
    \end{tabular}%
}

\label{tab:result_table_2}

\end{table*}

Table \ref{tab:result_table} presents Recall@1/5 comparison against state-of-the-art VPR methods on standard outdoor street-view datasets, which are similar to the typical training datasets used for VPR~\cite{arandjelovic2016netvlad,berton2022rethinking,ali2023mixvpr}. Table~\ref{tab:result_table_2} covers `out-of-distribution' datasets, inspired by AnyLoc~\cite{keetha2023anyloc}, covering indoor environments (Baidu Mall and 17 Places), aerial imagery (VPAir), indoor-to-outdoor viewing (InsideOut), and historical image matching (AmsterTime). Below, we discuss two key aspects of this comparative analysis: \textit{i)} how our segment-based approach compares against whole-image global descriptor based methods, and \textit{ii)} how performance trends vary depending on the choices of feature backbone with regards to task-specific (VPR) training.

\subsubsection{Aggregating Segments vs Whole Images}
Table~\ref{tab:result_table} and Table~\ref{tab:result_table_2} show that our proposed method \ourM{} achieves a new state-of-the-art on the majority of datasets, considering both the backbone variants: PreT and FineT. AnyLoc and SALAD respectively differ from \ourM{}-PreT and \ourM{}-FineT in terms of the aggregation scope (global vs segments). Thus, the superior performance of \ourM{} clearly highlights the role of segments based retrieval over whole-image based approach. On the Baidu Mall dataset -- highly-aliased indoor environment -- our method (pre-trained) improves over AnyLoc by $3-5\%$ for R@1 and around $6\%$ for R@5 in absolute gains. On the InsideOut dataset -- matching outdoor images viewed from within indoors -- our method leads to a `meaningful' recall, unlike all other baselines. Overall, these results highlight that even with the use of powerful image encoders (DINOv2), global aggregation struggles to deal with the challenges of matching images across major viewpoint shifts -- it is thus the partial image representation and matching which is needed to obtain superior recognition performance.
  
\subsubsection{VPR Fine-tuned Encoders + Segments}
For \ourM{}-FineT, we used a DINOv2 backbone finetuned for the purpose of VPR, mainly to observe the benefit of segments over global descriptor based approach in a \textit{task-specific manner}. Table~\ref{tab:result_table} and Table~\ref{tab:result_table_2} show that, on the outdoor street-view datasets, SALAD (finetuned DINOv2) generally performs better than AnyLoc (its pretrained counterpart), whereas the latter generally outperforms the former on `out-of-distribution' datasets. It can be clearly observed that these performance patterns translate well from global- to segment-level results.

\subsection{Revisiting Objects of Interest (OOI): Object Instance Retrieval}

A typical requirement of an embodied agent is to understand the context of its task through its memory/map information, which is composed of visual and/or semantic cues. For example, navigating to a given object goal requires a robot to visually recognize the goal and not be confused by perceptually-similar items. In this section, we demonstrate our method's ability to retrieve the correct image given just an Object Of Interest (OOI) as a query segment. For this purpose, we use an extended version of the Baidu dataset~\cite{weinzaepfel2019visual} which annotates OOI as various discriminative areas that can be reliably detected under variable viewpoint and lighting conditions. In total, there are 220 OOI, which cover various things such as logos, brand names, posters, etc., in a highly cluttered mall environment. To cast this dataset in terms of revisiting things, we use the original query images of the Baidu Mall~\cite{baidu_dataset_Sun2017ADF} dataset as the database and the images with OOI as the queries. This allows us to evaluate the OOIs directly. This is similar to VPR evaluation of recall in terms of image retrieval but with querying of a specific segment instead of using all the segments of the query image.  

\begin{table}[t]
    
    \centering
    \setlength{\tabcolsep}{4\tabcolsep}
    \caption{Recall@1 results for various approaches on Object-Instance Retrieval Task}
    \begin{tabular}{l|c|c|c|c}
    \hline

    \textbf{Method} & SegVLAD NoNbrAgg & SegVLAD & Segment-to-Global & Global-to-Global \\ \hline
    \textbf{R@1} & 64.1 & \textbf{92.7} & 30.0 & 86.4 \\ \hline
    \end{tabular}
    
    \vspace{3 pt}
    \label{tab:ooi_table}

\end{table}

We consider four different methods of recognizing known objects in this study. i) \texttt{Global-to-Global}: as a baseline method, we use whole images to represent and retrieve, i.e., without using the OOI mask; this resembles object-goal recognition problem for an InstanceImageNavigation task~\cite{krantz2022instance}. ii) \texttt{Segment-to-Global}: this is the same as the previous setting except that the query image descriptor is aggregated only using the OOI mask; this tests the ability of the image encoder/aggregator to match segment-level descriptor against global descriptors. iii) \texttt{\ourM{}} and iv) \texttt{\ourM{} NoNbrAgg}, which are our proposed methods but the latter does not use any neighborhood information; this highlights the relevance of spatial context around the OOI for recognition. For \ourM{}, we create a virtual segment mask for the OOI, append it to the other masks of the image, and then perform our neighborhood expansion and feature aggregation, as described in Section~\ref{sec:approach}.

Table \ref{tab:ooi_table} reports Recall@1 for different recognition methods. It can be observed that \texttt{\ourM{}} outperforms \texttt{Global-to-Global} matching by a large margin, which shows that recognizing specific object instances through their images (as in InstanceImageNav) is more prone to failures. It can further be observed from low recall of \texttt{\ourM{} NoNbrAgg} that neighborhood aggregation around segments is crucial to capture the required context. Finally, poor recall of \texttt{Segment-to-Global} highlights that matching a part of an image (OOI) with the whole image is not a viable solution for object instance recognition.  

\begin{table}[t]
    
\resizebox{\columnwidth}{!}{%
\begin{minipage}[t]{.45\linewidth}
    \centering
      \caption{Recall@1/5 for Baidu mall dataset for different aggregation methods and different orders of neighborhood expansion.}
    \resizebox{0.8\columnwidth}{!}{%
     \setlength{\tabcolsep}{5\tabcolsep}
    \begin{tabular}[t]{ccc}
    \toprule
        {\textbf{Order}} & {\textbf{SegVLAD}} &  {\textbf{SAP}}    \\ 
        \midrule
        0 &73.1/89.9   &  \textbf{ 74.6}/\textbf{91.1}  \\
        1 &77.4/91.7   &   65.6/87.2  \\
        2 &76.3/92.4   &   53.2/81.3  \\
        3 &\textbf{77.7}/\textbf{92.6 }  & 49.8/78.0   \\
        \bottomrule
    \end{tabular}%
    }
    \label{tab:recall_at_order}
    \end{minipage}%
    \hspace{0.01\linewidth}
\begin{minipage}[t]{.45\linewidth}
     \caption{Recall@1/5 comparison between different methods for ranking images based on segment-level retrieval.}
    \resizebox{0.9\columnwidth}{!}{%
    \setlength{\tabcolsep}{5\tabcolsep}
    \begin{tabular}[t]{lccc}
    \toprule
        {\textbf{Method}} & {\textbf{Baidu}}& {\textbf{AmsterTime}} \\
        \midrule
        Max Seg & \textbf{78.5}/\textbf{93.9} & 53.9/70.4 \\
        Max Sim & 65.2/92.7& 34.4/62.4 \\
       \textit{\textbf{Ours}} & \textbf{78.5}/93.8 & \textbf{54.4}/\textbf{76.3}\\
        \bottomrule
    \end{tabular}%
    }
    \label{tab:recall_methods}
    \end{minipage}
    }

\end{table}

\subsection{Ablation Studies}
\label{sec:ablate}
\subsubsection{Aggregation method \& Order of Neighborhood Expansion}
Previous studies on VPR such as AnyLoc~\cite{keetha2023anyloc} have shown VLAD to be better than other aggregation methods for whole-image based global descriptors. However, in an increasing number of segment-based approaches~\cite{gu2023conceptgraphs,chang2023goat,garg2024robohop,shlapentokh2024region}, segment \textit{average} pooling (SAP) is used more commonly. Thus, we compare hard-assignment VLAD against SAP on Baidu dataset. For SAP, we upsample the DINOv2 features to match the resolution of our SAM masks -- this upsampling is shown to enhance performance in~\cite{shlapentokh2024region}. For VLAD aggregation, we use our proposed method \ourM{}, where we downsample masks to match with the low resolution of DINOv2. Table~\ref{tab:recall_at_order} shows that SAP performs well for order 0 aggregation (i.e., no neighborhood aggregation) but its performance reduces as the neighborhood expands. On the other hand, \ourM{} has low recall when no neighborhood is considered but benefits significantly even with its immediate neighborhood (order 1). We attribute these inverted trends of \ourM{} and SAP to the very nature of these aggregation methods: as more information becomes available SAP smooths out the overall information content whereas \ourM{} benefits from additional information which gets distributed across its clusters, thus minimizing any possible smoothing effect. It can be observed that R@5 increases for \ourM{} with an increasing order of neighborhood expansion but margins diminish for higher orders. Overall, \ourM{} (order 3) achieves the best results, despite aggregating at a $14\times$ lower resolution than SAP's upsampling based aggregation.

\subsubsection{Segment to Image Retrieval} \label{sec:seg_ret}
Unlike conventional global descriptor retrieval based VPR, we perform retrieval for multiple SuperSegments of the query image. To obtain retrieval output in terms of images (as that is what VPR is typically evaluated on), there exist multiple ways to combine the top segment-level matches across all the query segments. We consider the following alternative options. i) \texttt{MaxSeg}: we obtain the best matching segment for each query segment, associate the matched segments to their respective reference image indices, and then rank these image indices based on their frequency; this method \textit{weakly} resembles an inverted index list based counting of common segments between the query and the reference. ii) \texttt{MaxSim}: across the best matching reference segments, we order their image indices based on the segment similarity; this method is similar to that used in MultiVLAD~\cite{arandjelovic2013all}. iii) \texttt{Similarity-Weighted Frequency}: this is our proposed method as defined in Section~\ref{sec:app_seg}.
Table~\ref{tab:recall_methods} shows that our proposed method for combining segment-level hypotheses consistently achieves superior results for both the datasets. While MaxSeg achieves a similar performance on Baidu, it suffers a drop in recall for AmsterTime. Both the methods outperform MaxSim at R@1 by a large margin.

\def\scaleQS{0.3}
\begin{figure}[t]
    \centering
        \setlength{\tabcolsep}{2\tabcolsep} 

    \begin{tabular}{ccc}
         \includegraphics[width=\scaleQS\linewidth]{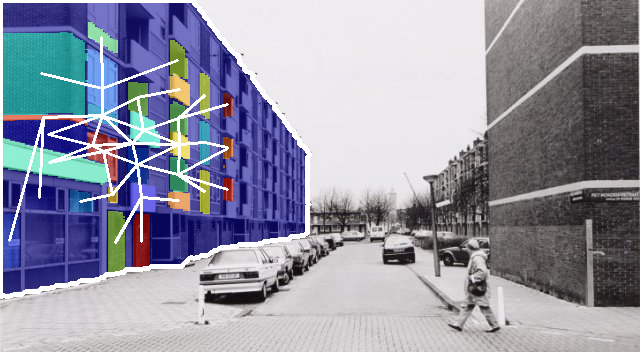}& \includegraphics[width=\scaleQS\linewidth]{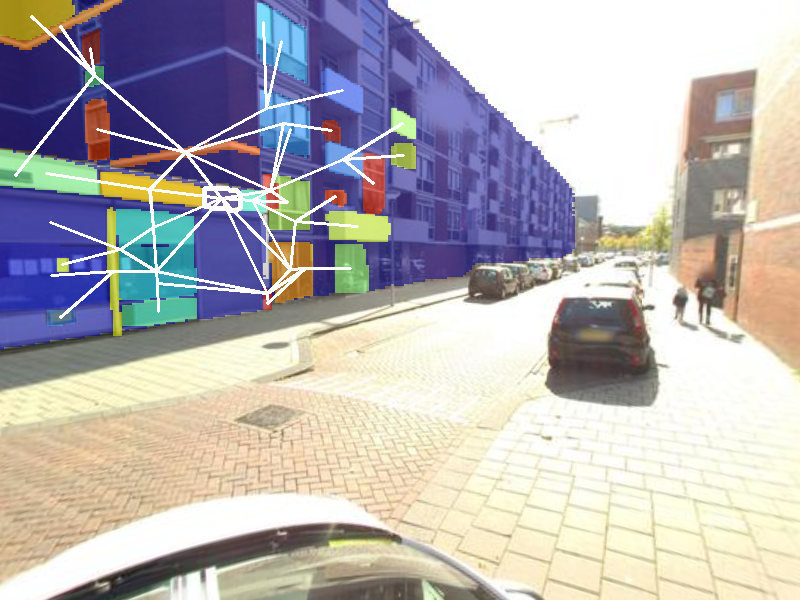} & \includegraphics[width=\scaleQS\linewidth]{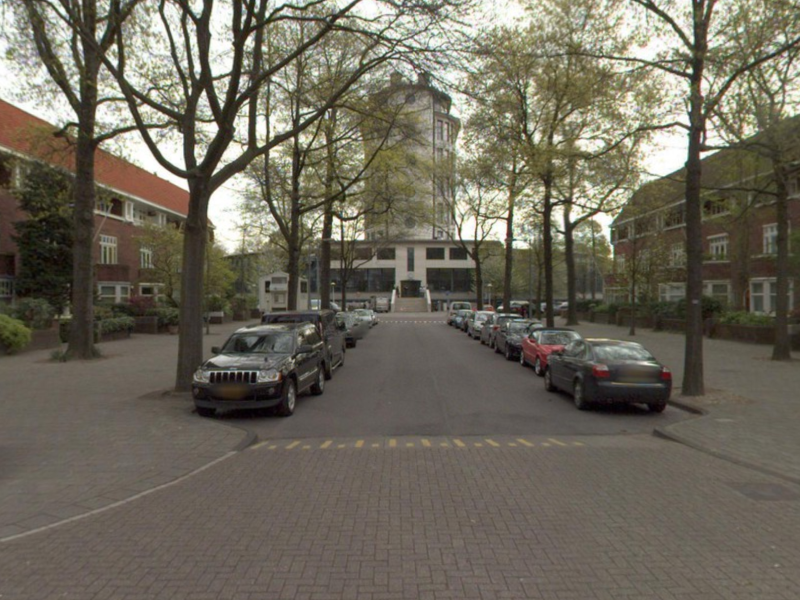} \\
         \includegraphics[width=\scaleQS\linewidth]{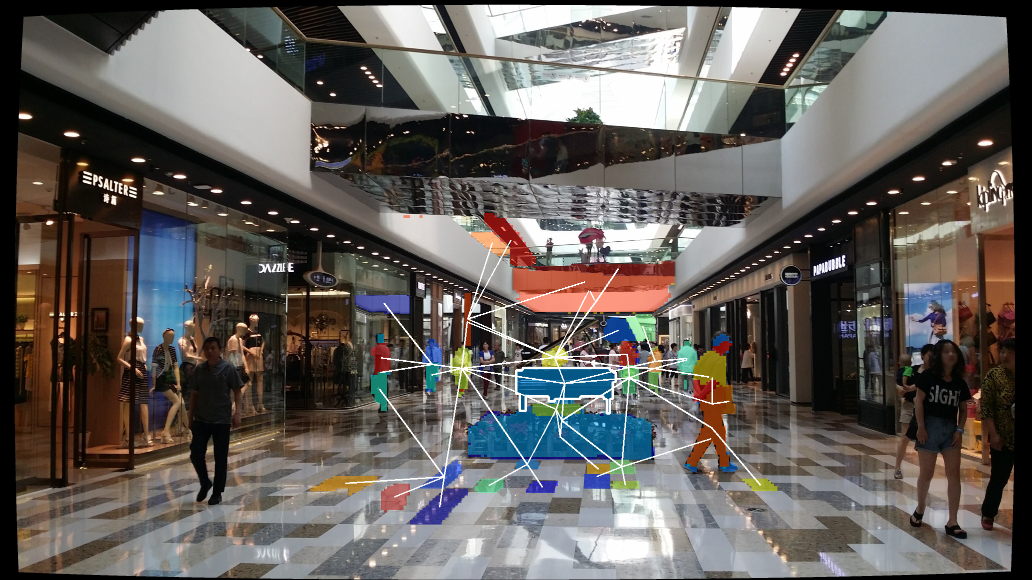}& \includegraphics[width=\scaleQS\linewidth]{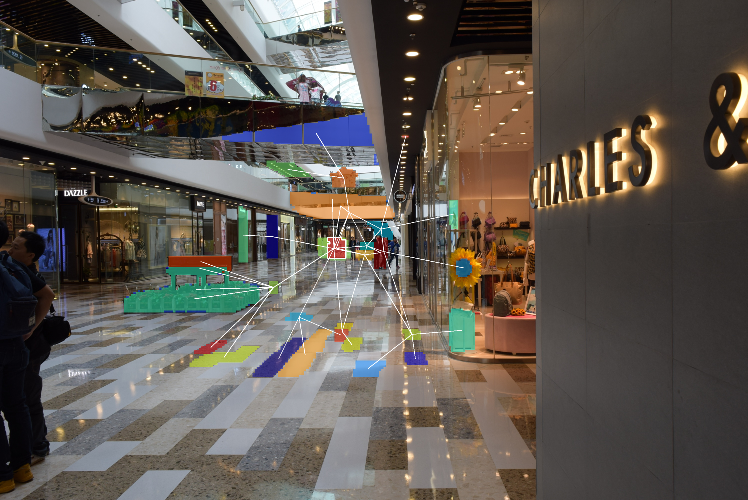} & \includegraphics[width=\scaleQS\linewidth]{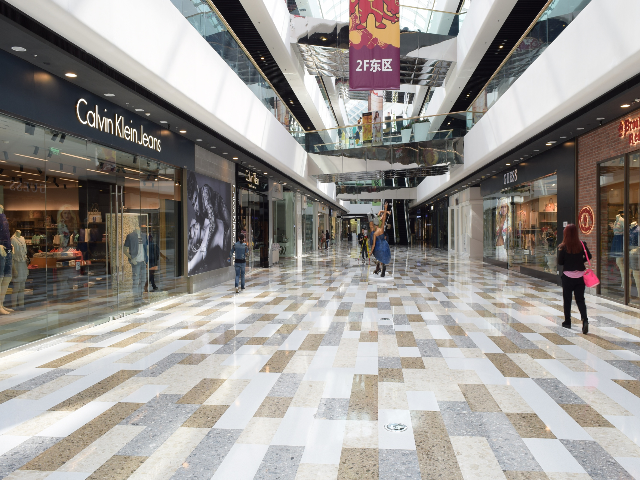} \\
         \includegraphics[width=\scaleQS\linewidth]{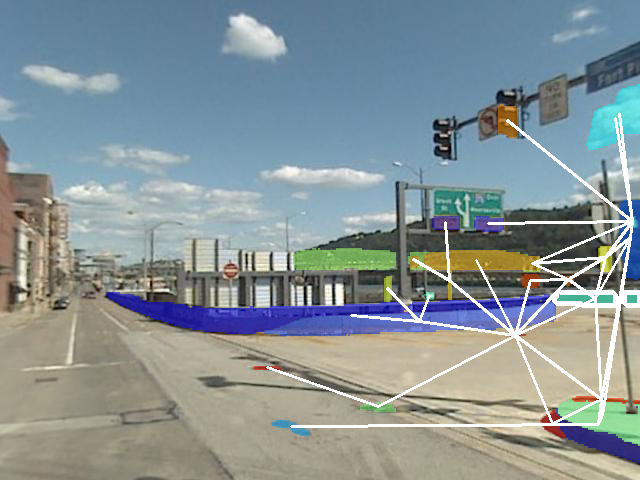}& \includegraphics[width=\scaleQS\linewidth]{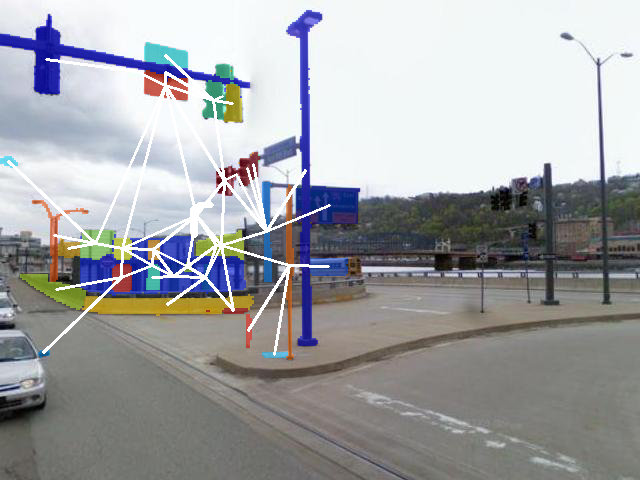} & \includegraphics[width=\scaleQS\linewidth]{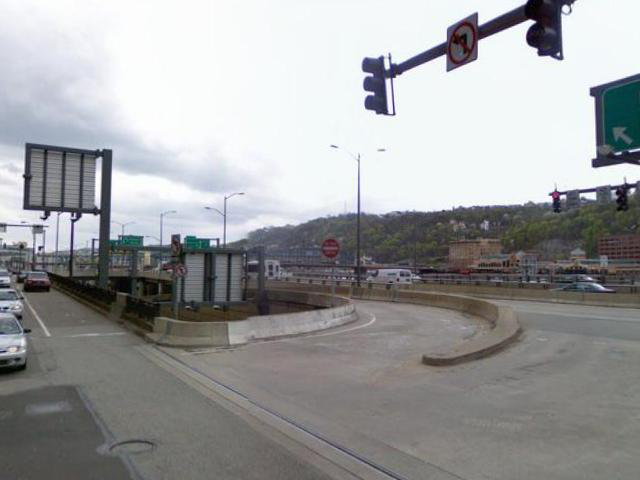} \\
    \end{tabular}%
    \caption{Qualitative results: Columns respectively represent the query image, correct match of \ourM{} and incorrect match of AnyLoc. Examples from different datasets: AmsterTime, Baidu Mall, Pitts-30K are presented across the rows. }
    \label{fig:qual}
\end{figure}

\subsubsection{Patches vs Segment}\label{sec:patches_vs_segments}
\begin{figure}
    \centering
    \begin{minipage}{0.48\textwidth}
        \centering
        \includegraphics[width=0.9\textwidth]{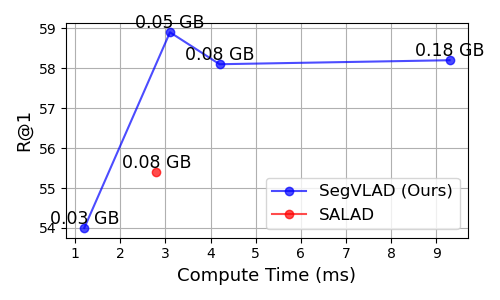}
        \caption{Recall vs storage/retrieval time on AmsterTime.}
        \label{fig:compute_graph}
    \end{minipage}%
    \hfill
    \begin{minipage}{0.48\textwidth}
        \centering
        \captionof{table}{Patch vs Segments on AmsterTime (Reso. 256$\times$256)}
    \resizebox{0.9\columnwidth}{!}{%
        \begin{tabular}{cccc}
        \toprule
        \textbf{{Patch Size}} & \textbf{NumSegDb}  & \textbf{NumSegIm} & {\textbf{R@1/5}} \\
        \midrule
        16$\times$16 &   315136  & 256   & 35.7/62.1 \\
        32$\times$32 &   ~78784  & ~64   & 45.9/72.5 \\
        64$\times$64 &  ~19696  & ~~16   & 52.0/75.1\\
        128$\times$128 & ~~~4924 & ~~~4 & 53.5/65.6 \\
        \midrule
        SegVLAD &  129637   & 105    & \textbf{56.8}/\textbf{77.7}\\
        \bottomrule
        \end{tabular}
        }
        \label{tab:patch_vs_seg_exp}
    \end{minipage}
\end{figure}

While open-set segmentation~\cite{kirillov2023segment} is aimed at a meaningful segregation of visual entities, a simple alternative to our segment-based retrieval is to use uniformly defined regions/patches. Table~\ref{tab:patch_vs_seg_exp} compares SegVLAD with a patch-based approach, where we consider arbitrary square patch sizes to segment an image. It can be observed that SegVLAD outperforms its patch-based counterparts, where smaller patches perform the worst, and for larger patches, R@1 saturates while R@5 reduces. These results are also in line with similar findings of a recent work~\cite{shlapentokh2024region}.

\subsection{Limitations, Storage, Compute Time \& IOU Based Filtering} \label{sec:iou}
A key limitation of our method is its large map size, i.e., large storage requirement for segment-level descriptors (see supplementary for further details).
In this section, we analyze the resource requirements in terms of database (index) storage and query retrieval time for our method, along with a preliminary study on IOU (Intersection over Union) based filtering of SuperSegments to reduce such costs. We compute IOU between all pairs of SuperSegments in a given image, and remove segments with $IOU(s_i,s_j) > \psi \; \forall \; i\in [1,S], \, j \in [i,S]$, where $\psi \in [0,1]$ is a threshold on IOU and $s_i$ refers to the list of SuperSegments sorted by their pixel area in a descending order. We only perform this culling on the database segments.
We use the outdoor-finetuned DINOv2 backbone for this purpose and compare SegVLAD with SALAD on AmsterTime (see supplementary for additional results on Pitts30K). Figure~\ref{fig:compute_graph} shows that SegVLAD outperforms SALAD while requiring less storage (annotated on points) and comparable retrieval time (excludes extraction time), using IoU-based filtering threshold ranging from $0.2$ to $0.8$ (left to right) with a step size of $0.2$. In particular, at $0.4$ IOU threshold, only $20\%$ of SuperSegments are retained ($0.05$ GB) while still outperforming the baseline.

\subsection{Qualitative Analyses}
In this section, we further demonstrate the capabilities of our method through qualitative visualizations. We compare our method against AnyLoc, where the only difference between the two methods is \textit{Global} aggregation/retrieval vs \textit{SuperSegment} aggregation/retrieval. We particularly consider the queries for which our approach successfully retrieved the correct match but AnyLoc failed to do so (additional examples can be found in the supplementary).
Figure~\ref{fig:qual} shows triplets of images in the order of query, correct match (ours), and incorrect match (Anyloc). The segmented part shows one of the correctly matched SuperSegments, displayed as a subgraph in white color overlaid on the corresponding segment masks. 

The first row shows a triplet from AmsterTime where our proposed method is able to correctly recognize a subgraph of building across the image pair, whereas Anyloc retrieves an incorrect image of a street-view with buildings, cars, and road, laid out similarly across the image pair. This highlights that a global descriptor can get confused with \textit{perceptually-aliased global context}. In contrast, our SuperSegment based \ourM{} is not only able to retrieve the correct image but it also correctly finds the mutually-overlapping area. A similar trend follows for the Baidu Mall (middle row) and Pitts30K (last row). In the Baidu Mall example, our approach identifies the piano and the region around it in the query image. It correctly retrieves an image having similar spatial context with the piano. This is akin to how humans use spatial context to identify places. AnyLoc, on the other hand, retrieves an image with similar floor tiles and railings. This example particularly highlights our hypothesis that dissimilarity of non-overlapping regions can dominate the similarity of overlapping regions in global whole-image descriptors. Finally, the Pitts30K example (last row) shows a case of strong viewpoint change. While \ourM{} correctly matches the traffic signal and signboards to retrieve the correct match, AnyLoc retrieves a similar looking image while missing the finer details. This example particularly reinforces the idea of `revisiting things', as even though the background mountain is common across the triplet, it is the context of the things near the camera/robot which helps in uniquely recognizing a place.

\section{Conclusion and Future Work}
In this paper, we presented a novel visual place recognition pipeline \textit{\ourM{}} based on image segments-based description and retrieval, which is akin to `revisting things' as a means to recognize specific instances of what constitute a place. Our proposed \textit{SuperSegments} based image representation and a novel factorization based feature aggregation enables us to effectively represent and retrieve images using our segment similarity-weighted image ranking. Our results show that despite using powerful image encoders such as DINOv2 (pretrained or VPR-finetuned), existing global descriptor based techniques are unable to deal with the challenges of viewpoint variations. In contrast, \ourM{} is able to correctly retrieve images through its ability to match partially-overlapping images with its partial image representation in the form of semi-global subgraphs of segments, i.e., SuperSegments. Thus, our method achieves state-of-the-art results on three diverse datasets (indoor and outdoor) that exhibit strong viewpoint variations on top of other challenges of appearance shift and high perceptual aliasing. Through an additional object instance retrieval study, we  demonstrate the unique ability of our method to recognize objects instances within their specific place contexts -- an open-set recognition capability that existing VPR methods lack.

Our approach shifts the paradigm in retrieval based VPR research, as the conventional methods predominantly classify into either whole-image global descriptor based coarse retrieval or local feature based geometric reranking. Our approach complements recent concurrent works like MESA~\cite{zhang2024mesa}; future work can explore a hierarchical VPR pipeline that closely integrates a segment-based coarse retriever with segments-based reranker such as MESA, thus doing away entirely with global whole-image descriptors. Furthermore, segments-based representation with implicitly baked semantics provide a natural way for creating \textit{text-based} interfaces through CLIP~\cite{radford2021learning} and LLMs (Large Language Models)~\cite{brown2020language}, which can be easily integrated with recent efforts in this direction~\cite{zou2024segment,maalouf2024follow,folorunsho2024semantic,garg2024robohop,manglani2023real,chang2023goat,gu2023conceptgraphs}.

\clearpage  %

\chapter*{Supplementary Material}

In this supplementary, we first present the limitations of our work, coupled with additional results on IOU-based filtering of SuperSegments (\cref{sec:supp_iou}). This is followed by ablation studies on local feature based retriever (\cref{sec:lfr}) and an efficient version of SAM (\cref{sec:FastSegVLAD}). We then present implementation and benchmarking details relating to the proposed factorized feature aggregation method (\cref{sec: agg_math}), backbone models (SAM and DINOv2) (\cref{sec:backbones}), and benchmark datasets (\cref{sec:datasets}). Finally, we present additional qualitative examples for retrieval (\cref{sec:qual}).

\section{Limitations}
While our proposed method \ourM{} achieves state-of-the-art results on a diverse set of VPR benchmark datasets, there are notable limitations of our approach. i) Redundancy: we create several overlapping SuperSegments per image. While these are somewhat necessary to enable accurate partial image matching via partial representations, SuperSegments formed through neighboring central segments will have a very high overlap. ii) Map size, we need to store several SuperSegment descriptors (far more than a typical global descriptors database). These limitations to some extent can be addressed through simple measures, e.g., masks IOU based filtering of the database segments to reduce both the redundancy and storage, as detailed in the subsequent section. The value of our research primarily lies in the demonstration of a novel approach to VPR that not only addresses the fundamental limitation of global descriptors but is also characterized as an open-set, object-based, text-interface-friendly representation, which is more likely to plug in to similar recent approaches aimed at embodied intelligence.

\section{Storage, Compute Time \& IOU Based Filtering Additional Results} \label{sec:supp_iou}

Table~\ref{tab:iou_table_at} shows results for IOU-based filtering for different thresholds, corresponding number of database segments, their storage consumption, and the average retrieval time per query. As can be observed from the results for AmsterTime, 0.4 IOU threshold removes up to $80\%$ of the original segments, and our method still outperforms the global descriptor baseline while only requiring roughly half its storage. For Pitts30K, both 0.4 and 0.6 IOU thresholds remain reasonable choices, with improved recall at reduced storage and time. Furthermore, in absolute terms, both storage and retrieval time
for our method are practically viable, and comparable to the global descriptor baseline.

\begin{table}[t]
\caption{Storage (GB) and Retrieval Time (ms) analysis coupled with IOU based filtering of SuperSegments, compared to typical global descriptor based retrieval pipeline. Both global and segment based approaches use finetuned DINOv2 encoder.}
    \centering
    \resizebox{1\columnwidth}{!}{%
    \setlength{\tabcolsep}{3\tabcolsep}
    
    \begin{tabular}{lcccccccccc}
    \toprule
    &&& \multicolumn{4}{c}{AmsterTime} & \multicolumn{4}{c}{Pitts30K} \\
    \cmidrule(lr{0.75em}){4-7}
    \cmidrule(lr{0.75em}){8-11}
        {Method} & {$\psi$} & {Dim}
     &  $N_{Db}$ &  R@1/5 & Storage & Time & $N_{Db}$ & R@1/5& Storage & Time  \\ 
    \cmidrule(lr{0.75em}){1-3}
    \cmidrule(lr{0.75em}){4-7}
    \cmidrule(lr{0.75em}){8-11}

        SALAD  & -  & 8448& 1231 & 55.4/75.6 & 0.08& 2.8 & 10000  & 92.6/96.5 & 0.62 & 25.1 \\
    
    SegVLAD & 0.2 &1024& 3886 & 54.0/69.2 & 0.03 & 1.2  & 25704 &91.8/96.2 & 0.19 & 8.0\\  
    
   \textbf{SegVLAD} &  \textbf{0.4} & \textbf{1024}& \textbf{6200} & \textbf{58.9/76.2} &\textbf{ 0.05} & \textbf{3.1} & \textbf{40507} & \textbf{92.6/96.7} &\textbf{ 0.31} & \textbf{12.3}\\
    
    SegVLAD & 0.6 &1024& 9986 &58.1/77.3 & 0.08 & 4.2 & 65699 &92.8/96.8 & 0.51 & 19.2\\
    
    SegVLAD & 0.8 &1024& 23807 &58.2/79.5 & 0.18 & 9.3  & 154854 &92.4/96.8 & 1.18 & 43.2\\ 
    
    SegVLAD & 1.0 &1024& 129637 &58.9/79.3 & 0.98& 42.3 & 871150 &93.2/96.8 & 6.65& 251.1\\ 

    \bottomrule
    \end{tabular}%
}
    \label{tab:iou_table_at}
    
\end{table}

\section{Additional Results and Ablation Studies}
This section describes two ablation studies: \textit{i)} local feature based retrieval, which highlights the role of SuperSegments in capturing sufficient pixel scope to retrieve correctly, and \textit{ii)} efficient version of SAM, which emphasizes the lack of strict dependence on a particular segmentation model while also addressing its computational bottleneck.

\subsection{Local Feature Retrieval} \label{sec:lfr}
There exist retrieval-based place recognition methods that directly use the local features with efficient inverted indexing and searching, e.g., DeLF~\cite{noh2017large}. In this section, we compare our segments-based retriever against such a local feature-based retriever, while using the same feature backbone (AnyLoc's DINOv2). We sample $S$ local features uniformly at random, where $S$ is the average number of segments for that image. As we intend to compare the role of local features in contrast to aggregation at segment/global level, we directly construct a flat index of these local features using the reference images of the given dataset. We then use our retrieval pipeline, considering local features a drop-in replacement of segment descriptors.

Table~\ref{tab:local} compares recall across local (pixel/point) features, segment descriptors and global descriptors on Baidu (indoor) and AmsterTime (outdoor). $S$ is set respectively to $130$ and $100$ for the two datasets. It can be observed that local features perform inferior to both segments and global descriptors. This showcases that without any locally-aggregated information \textit{local features lack sufficient spatial context} needed to differentiate between two places. On the other hand, \textit{global descriptors carry additional spatial context} due to the non-overlapping parts of the image pair, which leads to mismatches. As a middle-order aggregation approach, our SuperSegments based \ourM{} descriptors achieve the required balance between aggregating sufficient spatial context and maintaining the distinctiveness necessary for viewpoint-invariant VPR task.

\begin{table}[t]
\caption{Comparison of local, segments and global retrieval methods using AnyLoc's DINOv2 encoder.}     
\centering
\setlength{\tabcolsep}{2.75\tabcolsep}
    \centering
    \begin{tabular}{lcc}
   
   \toprule

        \multirow{2}{*}{Method} & Baidu  & AmsterTime  \\
       
         &R@1/5 &R@1/5 \\
      \midrule
        {Local (Pixel/Point)} & 69.1/88.2  &42.2/66.1 \\
        {Segment (\ourM{})} & \textbf{78.5}/\textbf{93.8}  &\textbf{56.8}/\textbf{77.7}\\
        Global (AnyLoc) & 75.2/87.6 & 50.3/73.0 \\
   \bottomrule
   \end{tabular}%

\label{tab:local}
\end{table}
\begin{table}[t]
    
\caption{Comparing R@1/5 and segmentation inference time on the Baidu dataset using different segmentation methods: SAM vs FastSAM.}

\centering
\setlength{\tabcolsep}{0.75\tabcolsep}

  \begin{tabular}{lccccc}
 \toprule
      {Method }& {Scope} & {Backbone}~~ &   {Seg. Time (s)}~~~ & Resolution  & {R@1/5} \\
      \midrule
    AnyLoc~\cite{keetha2023anyloc}  & Global & DINOv2  & - & - & {75.2}/87.6 \\
      \midrule
    {\ourM{}} with SAM  & Segment  & DINOv2 & 3.5  & 240$\times$320  & \textbf{78.5}/\textbf{93.8} \\
    {\ourM{}} with FastSAM ~~~~ & Segment  & DINOv2 & 0.28  & 240$\times$320 & \textbf{76.2}/\textbf{91.9} \\
    {\ourM{}} with FastSAM ~~~~ & Segment  & DINOv2 &  0.32 & 480$\times$640  & \textbf{77.5}/\textbf{93.6} \\
 \bottomrule
 \end{tabular}%
\label{tab:result_table_Fast}

\end{table}

\subsection{SAM vs FastSAM}\label{sec:FastSegVLAD}

Section~\ref{sec:params} provides details of the configuration used for the original SAM model, where segment extraction becomes time consuming due to grid-based sampling/prompting (up to 3.5 seconds per image). Therefore, in our pipeline, we drop-in replace the original SAM with an efficient version of SAM, i.e., FastSAM~\cite{zhao2023fast} to analyse the recall-speed trade-off for our proposed method. FastSAM replaces the transformer architecture of SAM with a CNN-based detector trained on an instance segmentation task using 1/50th of the original training dataset of SAM. Additionally, it uses a prompt-free approach to attain high efficiency to \textit{segment everything}, which refers to extraction of all possible masks, as opposed to point/box based sparse prompting.

Table~\ref{tab:result_table_Fast} (Baidu) shows that the drop in recall values is quite marginal (up to $2\%$) when replacing SAM with FastSAM, while being $13\times$ faster in segmentation inference and still outperforming the second best method, i.e., AnyLoc (global descriptor baseline). In the last row, we further show that we can use FastSAM at higher input resolution with almost negligible increase in time and it further reduces the recall gap with original SAM.

\subsection{HardVLAD vs (Soft) NetVLAD}\label{sec:hard_soft_VLAD}
In the main paper, we used hard assignment based VLAD aggregation for both pretrained and finetuned DINOv2 backbones. The alternative choice, particularly for the finetuned version, is to use soft assignment as defined in NetVLAD~\cite{arandjelovic2016netvlad}. Table \ref{tab:hardVLAD_vs_softVLAD} shows that hard assignment of VLAD performs similar to NetVLAD.

\subsection{Different feature extractor}\label{sec:encoder_swap}
Other than the DINOv2 as our backbone feature extractor, here we consider another similar pretrained off-the-shelf backbone to highlight that our method is generally applicable to any dense feature extractor.
Table~\ref{tab:encoder_swap_diff} shows results for a more recent feature extractor, RADIOv2~\cite{ranzinger2024radio}. Our segments-based approach applied on top of this encoder substantially improves its performance, although our DINOv2-based SegVLAD achieves the best performance. 

\begin{table}[t]
    \centering
    \begin{minipage}[t]{0.48\textwidth}
        \centering
        \caption{R@1/5 Comparison between HardVLAD and SoftVLAD.}     

\centering
\resizebox{0.7\columnwidth}{!}{%
\setlength{\tabcolsep}{0.6\tabcolsep}
    \centering
    \begin{tabular}{lcc}
   
   \toprule
        {\textbf{Method}} & {\textbf{AmsterTime}}  &{\textbf{Pitts30K}} \\
        \midrule
        HardVLAD &  58.9/\textbf{79.3}  & \textbf{93.2}/96.8\\
        SoftVLAD &  \textbf{60.2}/78.8  &93/\textbf{96.9} \\
        
   \bottomrule
   \end{tabular}%
}

\label{tab:hardVLAD_vs_softVLAD}

    \end{minipage}
    \hfill
    \begin{minipage}[t]{0.48\textwidth}
        \centering
        \caption{Using Different feature extractor on AmsterTime.}     

\centering
\resizebox{\columnwidth}{!}{%
\setlength{\tabcolsep}{2\tabcolsep}
    \centering
    \begin{tabular}{lccc}
   
   \toprule
        {\textbf{Method}} & {\textbf{Backbone}}  &{\textbf{\#Params}} & {\textbf{R@1/5}} \\
        \midrule
        RADIO &  RADIOv2 ViT-H/16  & 0.6B   & 47.8/72.1 \\
        SegVLAD &  RADIOv2 ViT-H/16  & 0.6B   & \textit{52.9}/\textit{74.9}\\
        \midrule
        AnyLoc &   DINOv2 ViT-G/14  & 1.1B   & 50.3/73.0 \\
        SegVLAD &  DINOv2 ViT-G/14   & 1.1B    & \textbf{56.8}/\textbf{77.7}\\
   \bottomrule
   \end{tabular}%
}

\label{tab:encoder_swap_diff}

    \end{minipage}
\end{table}

\def\scaleRecall{0.45}
\begin{figure}[t]
    \centering
        \setlength{\tabcolsep}{1\tabcolsep} 
        \begin{tabular}{cc}
         \includegraphics[align=c, width=\scaleRecall\linewidth]{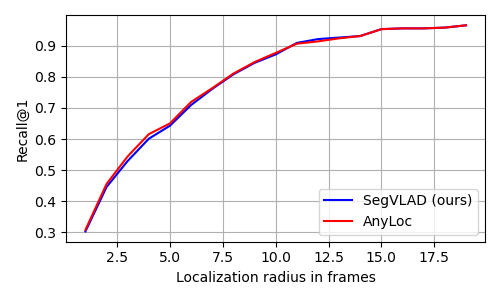} & 
            \includegraphics[align=c, width=\scaleRecall\linewidth]{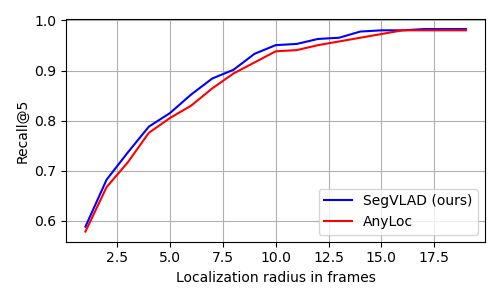}
            \end{tabular}%
          \caption{Recall vs localization radius (in terms of frame separation) for 17 Places. }
    \label{fig:17p_recall}
\end{figure}

\subsection{17Places dataset: Ground Truth Vagueness}
For VPR, ground truth is often defined in terms of either GPS coordinates~\cite{arandjelovic2016netvlad} or image frame separation~\cite{milford2012seqslam}. There is also often a discrepancy in defining ground truth for the VPR task, depending on how a `place' is defined and whether the camera position or visual overlap is used as a criterion for correct recognition, as discussed in~\cite{garg2021where}. In Figure~\ref{fig:17p_recall}, we present extended results for 17Places dataset to indicate how relative ranking of two methods can potentially switch depending on the choice of localization radius (in terms of frame separation).

\section{Implementation and Benchmarking Details} \label{sec:params}
In this section, we provide detailed implementation of our factorized representation for aggregation, backbone networks and benchmark datasets.

\subsection{Factorized Aggregation} \label{sec: agg_math}
We proposed a \textit{factorized} representation for feature aggregation as a unified method to aggregate at segment/global level for different aggregation types (see Section 3.3, Equation 2 in the main paper). In this section, we further elaborate this with explicit formulations for computing \textit{SegVLAD}, \textit{GlobalVLAD}, \textit{Global Average Pooling (GAP)}, \textit{Segment Average Pooling (SAP)}, and \textit{Generalized Mean Pooling (GeM)} using the proposed factorization: 
\begin{equation}
    F_{S\times D} = A_{S\times S}^o \cdot M_{S\times N}\ \cdot T_{N\times D}
    \label{eq:supp_agg}
\end{equation}
\paragraph{SegVLAD:} Given the segment adjacency matrix $A^o$ and binary masks $M$, \textit{each} cluster ($k$) undergoes segment-wise aggregation by setting $T^k$ as below: 
\begin{equation}
    T^k_{{N_k}\times D} = [T^k_1, T^k_2 \dots T^k_{N_k}]^\top; \quad T^k_p = \alpha_k(f_{p})(f_p-c_k)
    \label{eq:supp_vlad}
\end{equation}
For \textit{SegVLAD}, without neighborhood aggregation, $A^o$ is set as an identity matrix $I_S$.

\paragraph{GlobalVLAD:} $T$ is defined in the same way as Eq.~\ref{eq:supp_vlad}; segment adjacency matrix is not applicable and set to identity ($A^o = I_1$); and the segmentation mask is set as an all-ones matrix $M = J_{1\times N}$. Interestingly, the proposed factorization can further be adapted to obtain a \textit{GlobalVLAD} representation for all cluster centers in a \textit{single shot} by setting $M_{S\times N} = M'_{C\times N}$, where $M'$ represents membership of $f_p \; (p \in [1,N])$ in cluster $c_k \; (k \in [1,C])$.

\paragraph{Segment Average Pooling (SAP):} Given the segment adjacency matrix $A^o$ and binary masks $M$, the output of the image encoder is directly used as $T_{N\times D}$.

\paragraph{Global Average Pooling (GAP):} Similar to \textit{SAP}, $T_{N\times D}$ is the direct output of the image encoder, whereas the adjacency matrix is set to identity ($A^o = I_1$). The segmentation mask is set as an all-ones matrix $M = J_{1\times N}$, which is similar to \textit{GlobalVLAD}.

\paragraph{Generalized Mean Pooling (GeM):} The above formulations at both segments and global level can be easily extended to GeM~\cite{radenovic2018fine} through $T = T^p$ and $F = F^{1/p}$, where $p=1$ represents average pooling, $p=\infty$ represents max pooling and $p=3$ represents its typical use in previous works~\cite{berton2022rethinking,cao2020unifying,berton2023eigenplaces}.

\subsection{Backbone Networks} \label{sec:backbones}
\paragraph{DINOv2:} We follow AnyLoc~\cite{keetha2023anyloc} and use its default ViT-G backbone with the \texttt{value} facet features from layer $31$. For the DINOv2 \textit{finetuned} model, we followed SALAD~\cite{izquierdo2023optimal} which by default uses ViT-B backbone. Note that SALAD's aggregation is different from the soft assignment based VLAD aggregation proposed in NetVLAD~\cite{arandjelovic2016netvlad}. Since our method is based on Hard-VLAD aggregation, we followed SALAD's finetuning approach but replaced their aggregation with NetVLAD. Similar to SALAD, we only train the last 4 layers of DINOv2 (ViT-B) on the GSV dataset~\cite{ali2022gsv} with training image resolution as $224 \times 224$. Similar to NetVLAD, we used $64$ clusters which were initialized by randomly sampling images from the GSV training set.

\paragraph{SAM:} We use its ViT-H model with default parameters for segmentation. It generates masks for the entire image, using a grid of point prompts (32 along each edge), which are subsequently filtered based on IOU and stability score. 

\paragraph{Evaluation:}
For evaluation, we used $640\times480$ image resolution for DINOv2 encoder and $320\times240$ for SAM. For AmsterTime, we followed~\cite{AmsterTime_dataset_Yildiz_2022} and used a fixed resolution of $256\times256$ for both the models. Note that we follow the exact same procedure of image resizing when comparing our segments based approach with their global counterparts, i.e., AnyLoc~\cite{keetha2023anyloc} and SALAD~\cite{izquierdo2023optimal}. Additionally, for the baseline methods EigenPlaces~\cite{berton2023eigenplaces}, CosPlace~\cite{berton2022rethinking} and MixVPR~\cite{ali2023mixvpr}, we used ResNet50 backbone for all three methods with output descriptor dimensions of 2048 for EigenPlaces and Cosplace, and 4096 for MixVPR.

\subsection{Datasets} \label{sec:datasets}
In this work, we used a variety of datasets covering both outdoor and indoor environments. Outdoor datasets include Pitts30k~\cite{Pittsburgh-30k_dataset}, AmsterTime~\cite{AmsterTime_dataset_Yildiz_2022}, Mapillary Street Level Sequences (MSLS)~\cite{warburg2020mapillary}, SF-XL~\cite{berton2022rethinking}, Revisted Oxford5K and Revisited Paris6k~\cite{radenovic2018revisiting} . Indoor datasets include Baidu Mall~\cite{baidu_dataset_Sun2017ADF}, 17Places~\cite{zhou2017places} and InsideOut~\cite{ibrahimi2021inside}. While all these datasets exhibit strong viewpoint variations, they are significantly diverse in terms of appearance changes, perceptual aliasing (repetitive elements), type of environment/domain (indoor vs outdoor) and extent of temporal changes (e.g., matching historical images). We elaborate on each of the datasets in detail in the supplementary. In Table~\ref{tab:dataset_overview}, we report the number of reference and query images in each of these datasets. Furthermore, for our segments-based approach, we also note the total number of segments (based on SAM~\cite{kirillov2023segment}) across all the queries and references for each dataset, with the final columns listing the average number of segments per image. This highlights the scale at which we perform segments based retrieval, as opposed to global descriptors based retrieval.

\begin{table}[t]
\caption{An overview of the datasets in terms of the number of images (Ref/Qry), segments (Ref/Qry) and resolution used for image processing.}     

\centering
\resizebox{0.7\columnwidth}{!}{%
\setlength{\tabcolsep}{2.75\tabcolsep}
    \centering
    \begin{tabular}{lcccc}
   
   \toprule
        {Dataset} & {Num. Images}  &{ Num. Segments} & {Avg. Seg./Image} & Resolution \\
        \midrule
        Baidu &   ~~689  / 2292 & ~92K / 295K & ~~134 / 129 & 480$\times$640\\
        AmsterTime &  1231 / 1231 & 129K / 119K & 105 / 96 & 256$\times$256\\
        Pitts30K & 10K / 6816 & 873K / 592K & 87 / 86 & 480$\times$640\\
        MSLS CPH  & 6315 / 242 &  578k/23k  &  91/96  & 480$\times$640\\
        MSLS SF  &  12556/ 498  &  1.1M / 43k &  89/87  & 480$\times$640\\
        SF-XL (Val)  &  8015/7993  &  648k/ 646k &  81/ 81 & 512$\times$512\\
        InsideOut  & 10886 /500  &  821k/52k  &  75/106  & 480$\times$640\\
        17Places  &  406/406  &  34k/34k  &  84/86  & 480$\times$640\\
   \bottomrule
   \end{tabular}%
}

\label{tab:dataset_overview}

\end{table}

\def\scaleQS{0.18}
\begin{figure}[t]
    \centering
        \setlength{\tabcolsep}{1\tabcolsep} 
           
    \begin{tabular}{cccc}
        Query& \ourM{} (Ours)& AnyLoc~\cite{keetha2023anyloc} & SALAD~\cite{izquierdo2023optimal}\\
         \includegraphics[align=c, width=\scaleQS\linewidth, trim ={0.5cm 0.5cm 0.5cm 0.5cm}, clip]{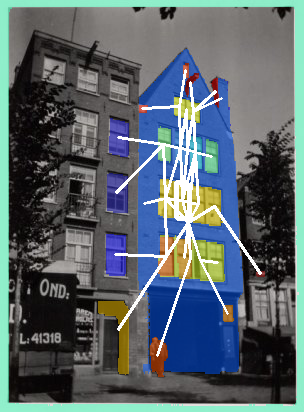}& \includegraphics[align=c,width=\scaleQS\linewidth]{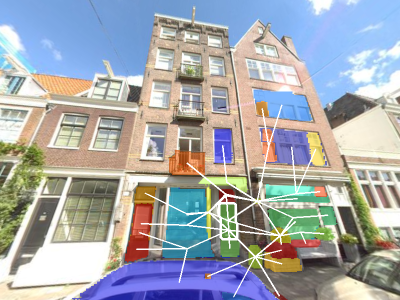} & \includegraphics[align=c,width=\scaleQS\linewidth]{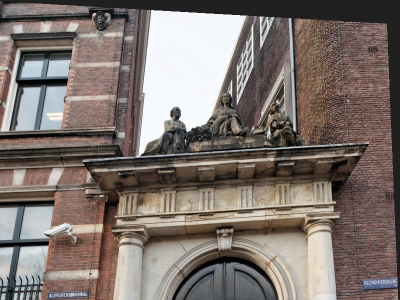} &
         \includegraphics[align=c,width=\scaleQS\linewidth]{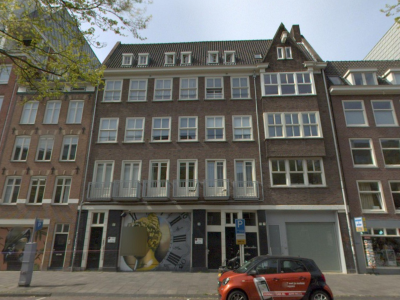}\\
         \includegraphics[align=c,width=\scaleQS\linewidth]{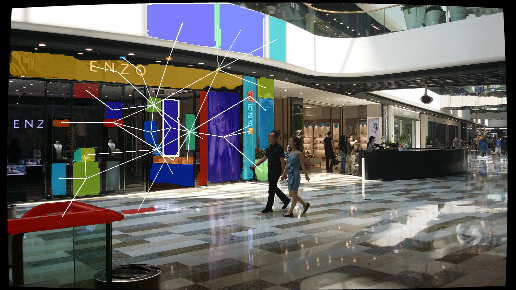}& \includegraphics[align=c,width=\scaleQS\linewidth]{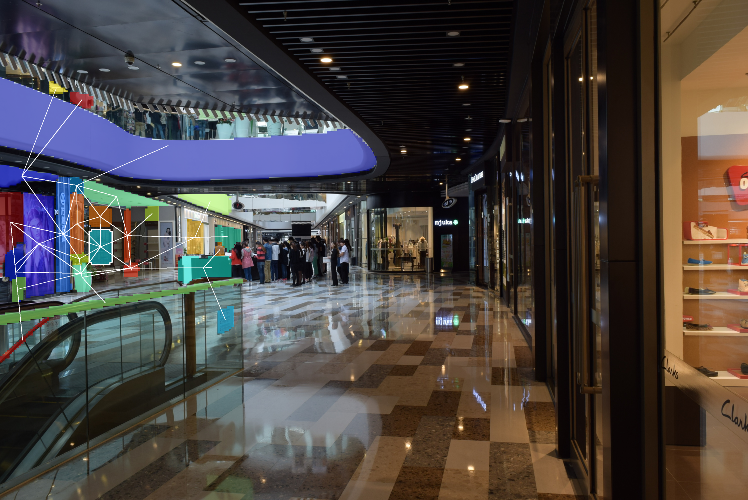} & \includegraphics[align=c,width=\scaleQS\linewidth]{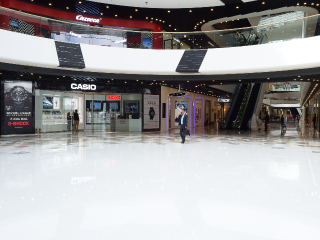} &
         \includegraphics[align=c,width=\scaleQS\linewidth]{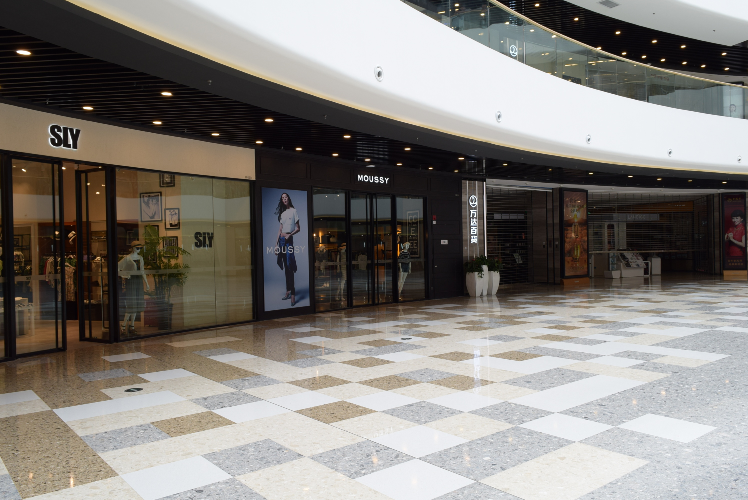}\\
         \includegraphics[align=c,width=\scaleQS\linewidth]{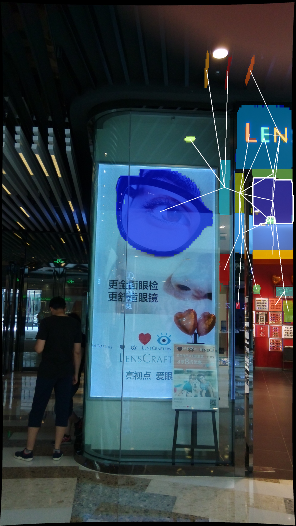}& \includegraphics[align=c,width=\scaleQS\linewidth]{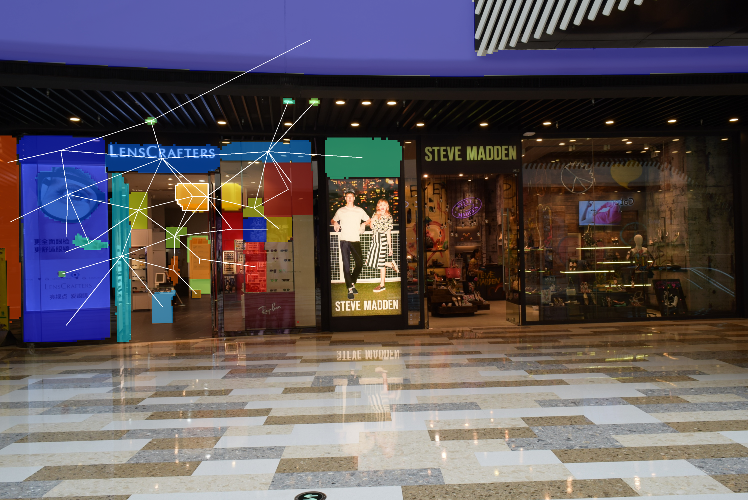} & \includegraphics[align=c,width=\scaleQS\linewidth]{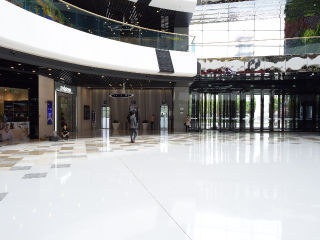}&
         \includegraphics[align=c,width=\scaleQS\linewidth]{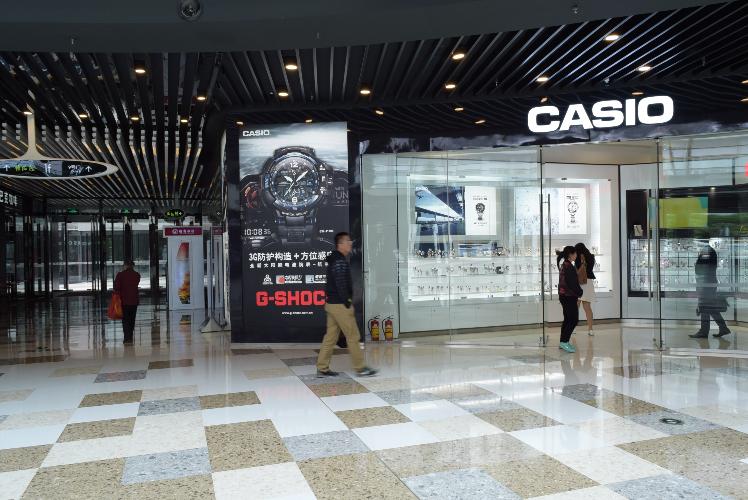}\\
         \includegraphics[align=c,width=\scaleQS\linewidth]{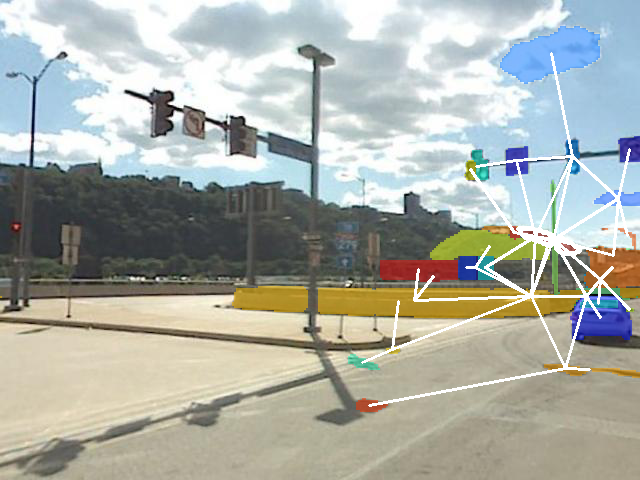}& \includegraphics[align=c,width=\scaleQS\linewidth]{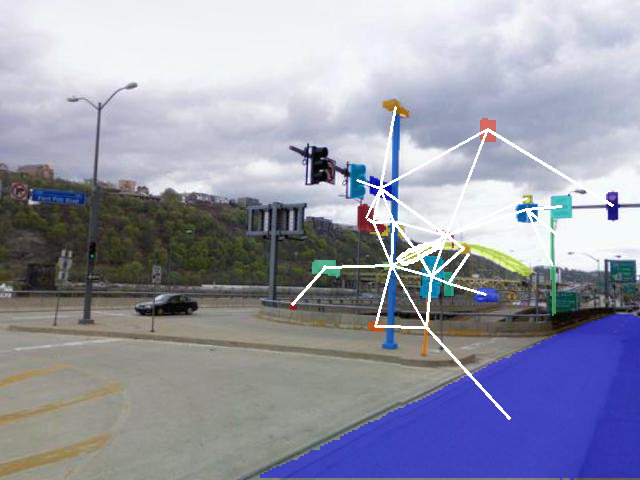} & \includegraphics[align=c,width=\scaleQS\linewidth]{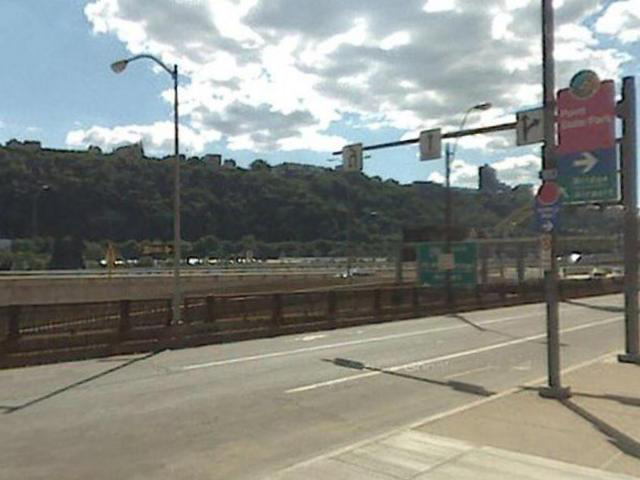}&
         \includegraphics[align=c,width=\scaleQS\linewidth]{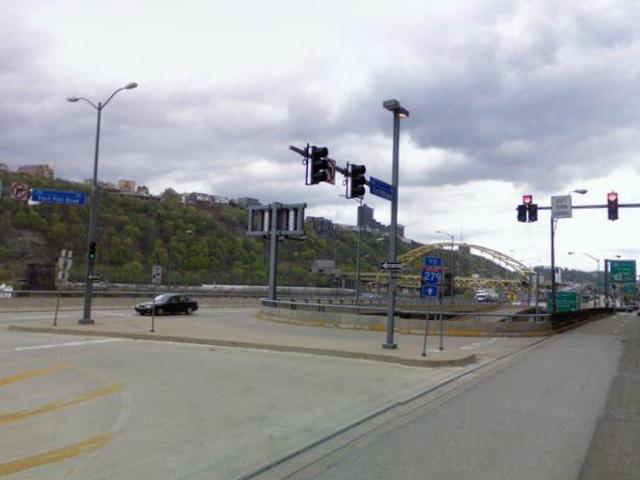}\\
         \includegraphics[align=c,width=\scaleQS\linewidth]{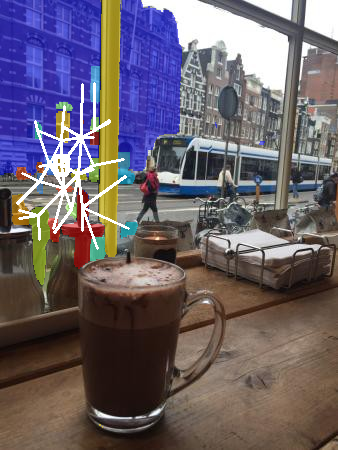} &
         \includegraphics[align=c,width=\scaleQS\linewidth]{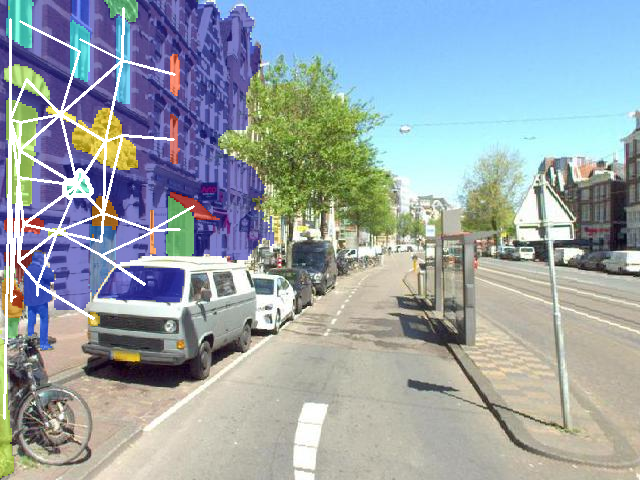} &
         \includegraphics[align=c,width=\scaleQS\linewidth]{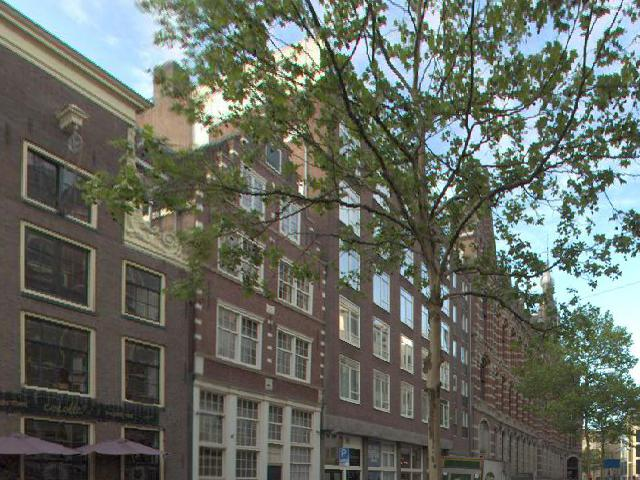} &
         \includegraphics[align=c,width=\scaleQS\linewidth]{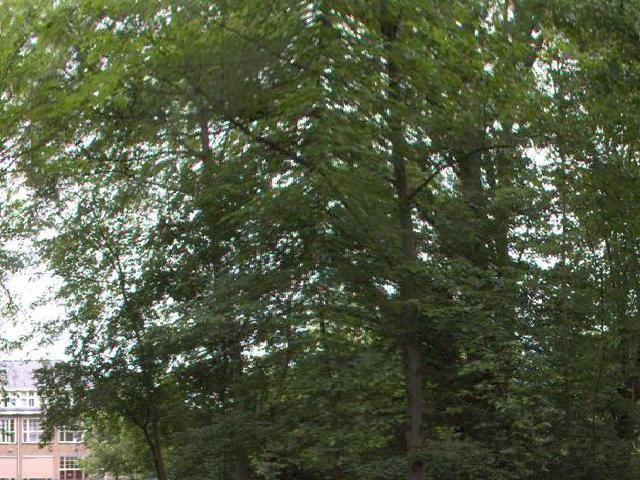} 
         \\
          \includegraphics[align=c,width=\scaleQS\linewidth]{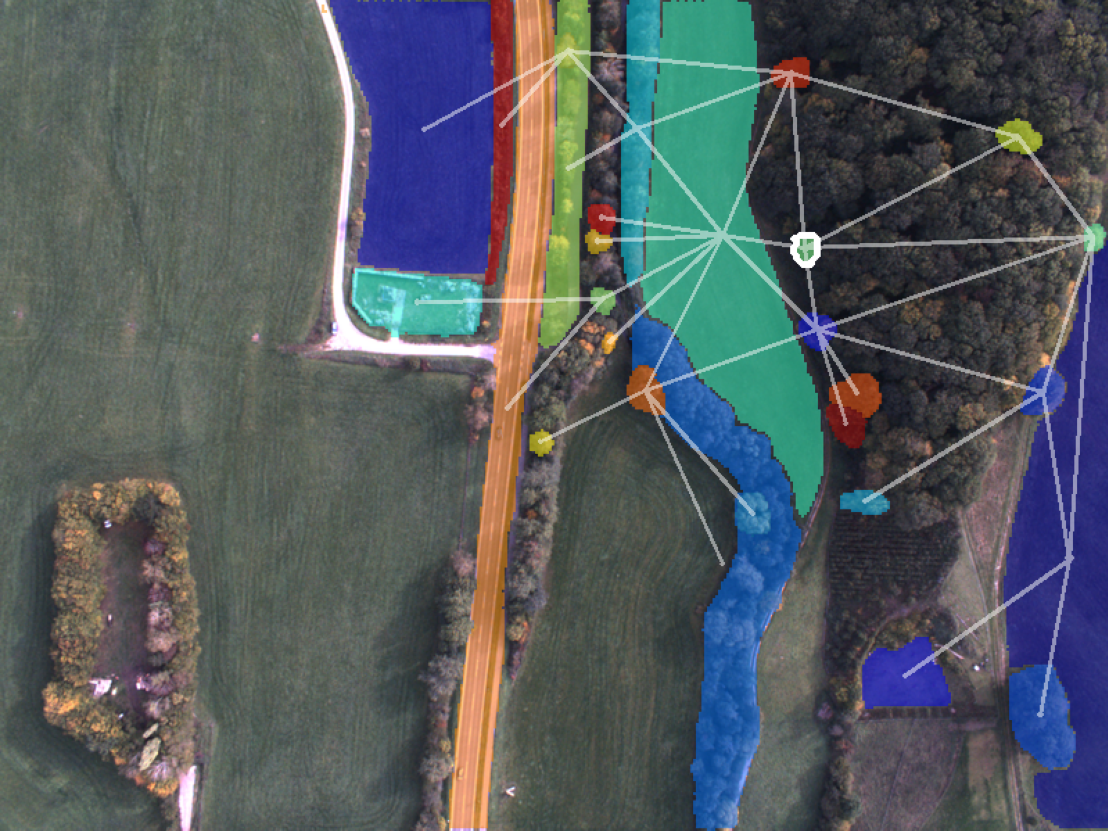} &
           \includegraphics[align=c,width=\scaleQS\linewidth]{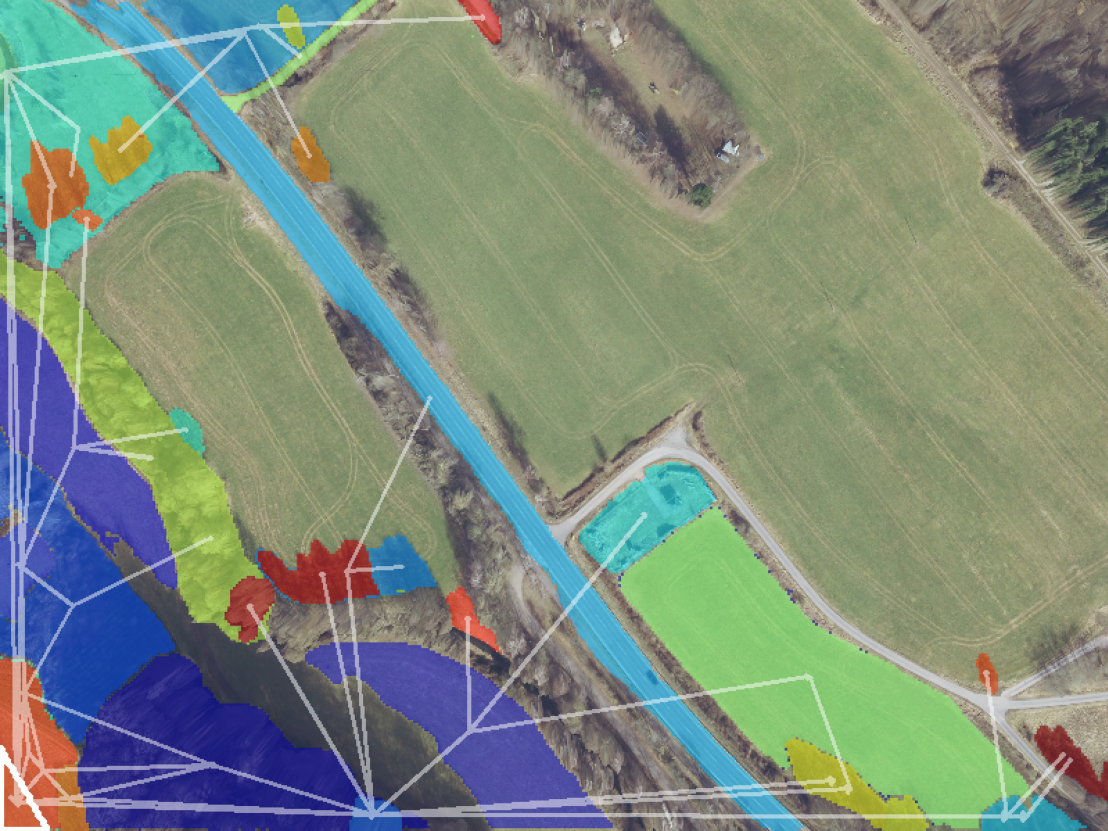} & \includegraphics[align=c,width=\scaleQS\linewidth]{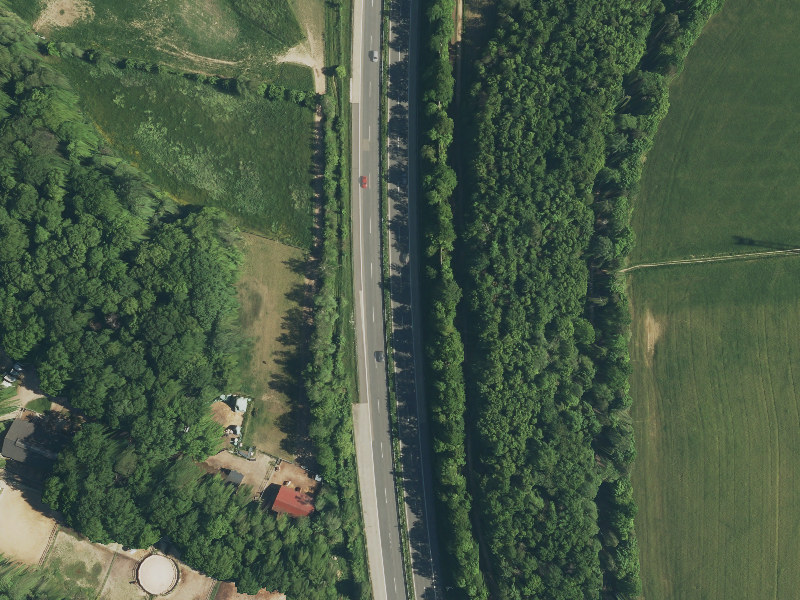} & \includegraphics[align=c,width=\scaleQS\linewidth]{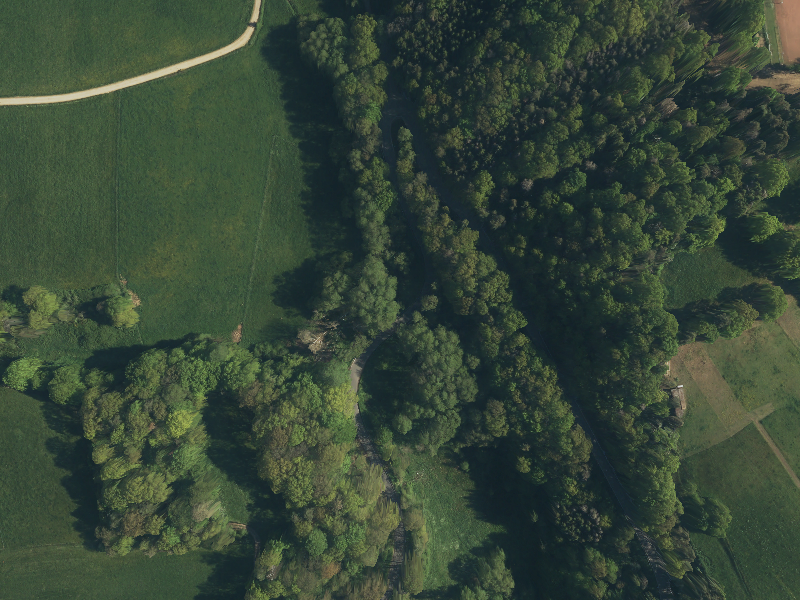} \\

\includegraphics[align=c,width=\scaleQS\linewidth]{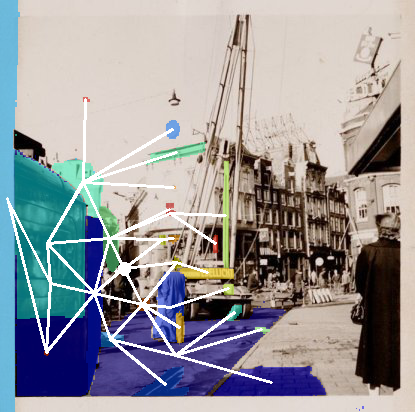}& \includegraphics[align=c,width=\scaleQS\linewidth]{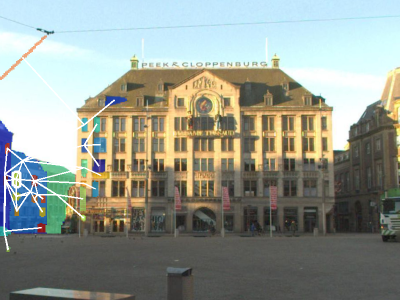} & \includegraphics[align=c,width=\scaleQS\linewidth]{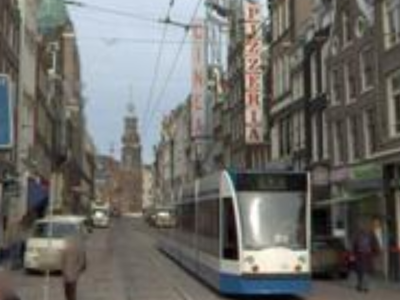}&
         \includegraphics[align=c,width=\scaleQS\linewidth]{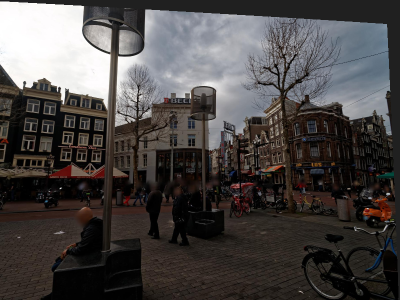}
         
    \end{tabular}%
    \caption{Qualitative results: Columns respectively represent the query image, and predictions from \ourM{}, AnyLoc, and SALAD. Examples from different datasets: AmsterTime, Baidu Mall, Pitts-30K, InsideOut and VPAir are presented across the rows. }
    \label{fig:supp_qual}
\end{figure}

\begin{enumerate}
    \item \textbf{Baidu Mall} dataset~\cite{baidu_dataset_Sun2017ADF} has images taken in a mall environment with varying camera poses. We use 2292 query images and 689 reference images. The dataset is characterized by extreme viewpoint shifts, cluttered environments, and semantic-rich information (logos, brand names, etc). It exhibits difficult appearance conditions of a commercial mall, characterized by reflective surfaces, non-uniform lighting, pedestrians, and perceptual aliasing caused by repetitive elements such as floor tiles, walls, stairs, and glass panes.
    \item \textbf{AmsterTime}~\cite{AmsterTime_dataset_Yildiz_2022} consists of 1231 pairs of reference and query images in the city of Amsterdam. It consists of grayscale historical images that are used as queries and modern RGB images of the same places (as confirmed by human experts) that are used as references. This dataset is characterized by long temporal changes, viewpoint shifts and modality (RGB vs grayscale). These drastic domain shifts make it a difficult dataset for VPR research.
    \item \textbf{Pitts30K}~\cite{Pittsburgh-30k_dataset} is one of the most used benchmarking datasets for VPR research. It consists of images taken from Google Street View showing different locations in downtown Pittsburgh with varying camera poses. We use the test split having 10000 references and 6816 query images. It is a large-scale dataset with several similar-appearing buildings and visual distractors like cars and pedestrians.
    \item \textbf{Mapillary Street-level Sequences}~\cite{warburg2020mapillary} (MSLS) is a large-scale, diverse dataset containing 1.6M street-level images from 30 cities worldwide. The dataset is divided into training (22 cities, 1.4M images), validation (2 cities, 30K images), and test (6 cities, 66K images) sets across different times of day, seasons, and new/old (after several years). This is especially useful as a non-saturated benchmark. We use the 2 cities from validation dataset, i.e. Copenhagen (CPH) and San Francisco (SF). The reference/query split of Copenhagen is 6315/242 whereas that of San Francisco is 12556/498.
    \item \textbf{SF-XL}~\cite{berton2022rethinking} This is a large scale dataset from the city of San Francisco with large-viewpoint changes. We use the validation split of this dataset with 8015 reference and 7993 query images.
    \item \textbf{17Places}~\cite{zhou2017places} This is an indoor dataset which consists of buildings at Coast Capri Hotel (British Columbia) as well as York University (Canada). Both reference and query datasets have 406 images. This dataset is challenging because of significant changes in lighting conditions as well as cluttered indoor environments. On this dataset particularly, we report results for both 5 (default) and 15 frames GT radius (r) as it incorrectly penalized correct retrieval for the former.
    \item \textbf{VPAir}~\cite{vpair_dataset_schleiss2022} This is an aerial dataset, consisting of 2706 database-query image pairs and 10,000 distractor images. These images are captured on an aircraft with a downward-facing camera at an altitude of 300 metres. Do note that we do not use the distractor set and our database and query images consist of 2706 images each, following AnyLoc's approach. Note that the dataset covers extremely challenging landscapes such as urban regions, farmlands and forests over more than 100 km and can be considered an out-of-distribution (OOD) dataset. Here, we use a localization radius of 3 frames.
    \item \textbf{InsideOut}~\cite{ibrahimi2021inside} This is a very interesting indoor/outdoor recognition dataset, wherein the task is to localize images based on outdoor scenes while viewing from indoor through windows. The images were taken in Amsterdam, the original dataset consisting of 6.4 million panoramic street-view images and 1000 user-generated indoor queries. However, we curate a smaller split of this dataset as follows: We use the test indoor queries set as our query images. For each query, we define a correct match if it is within a 50m radius; otherwise, it's classified as a distractor. To curate this reference set, we use 7 correct matches and 15 distractors per query image. After removing repetitive images the final dataset consists of 500 queries and 10886 reference images.
    \item \textbf{RO5k and RP6k}~\cite{radenovic2018revisiting} ROxford5K and RParis6k are classical instance retrieval benchmarks. Revisited Oxford5K has a reference/ query split of 4993/70 for 11 Oxford buildings and Paris6K has a reference query split of 6322/70 for 12 architectural groups. Each image of the same building is labelled as Good (i.e., positive), OK (i.e., positive), Junk, or Bad (i.e., negative) based on relevance. Junk images can be discarded or regarded as negative examples. 
\end{enumerate}

\section{More Qualitative Examples} \label{sec:qual}
In Figure~\ref{fig:supp_qual}, we present additional qualitative results for multiple datasets: Baidu, AmsterTime, Pitts30K, InsideOut, and VPAir, which not only highlight the extremities of viewpoint variations found in these datasets but also the capability of our method to retrieve correctly under such conditions. We compare \ourM{} with both AnyLoc and SALAD. The first column shows all the query images followed by the correct match predicted by \ourM{}, and incorrect matches predicted by AnyLoc and SALAD.
In the fourth row, we show an example from Pitts30K where \ourM{} and SALAD retrieve correctly but AnyLoc fails. The last row shows a case where all the methods fail.
We show all the qualitative results at the original resolution of the images.

In Figure~\ref{fig:ooi_qual}, we visualize the segments matched to the OOI queries using segment descriptors. While \ourM{} easily succeeds at recognizing large objects, it is particularly better than the global approach on recognizing small objects with the help of its expanded neighborhood. 

\def\scaleOOI{0.25}
\begin{figure}[t]
    \centering
    \setlength{\tabcolsep}{2\tabcolsep} 
    \begin{tabular}{cccc}
    \includegraphics[width=\scaleOOI\linewidth,trim={11cm 11cm 11cm 11cm},clip]{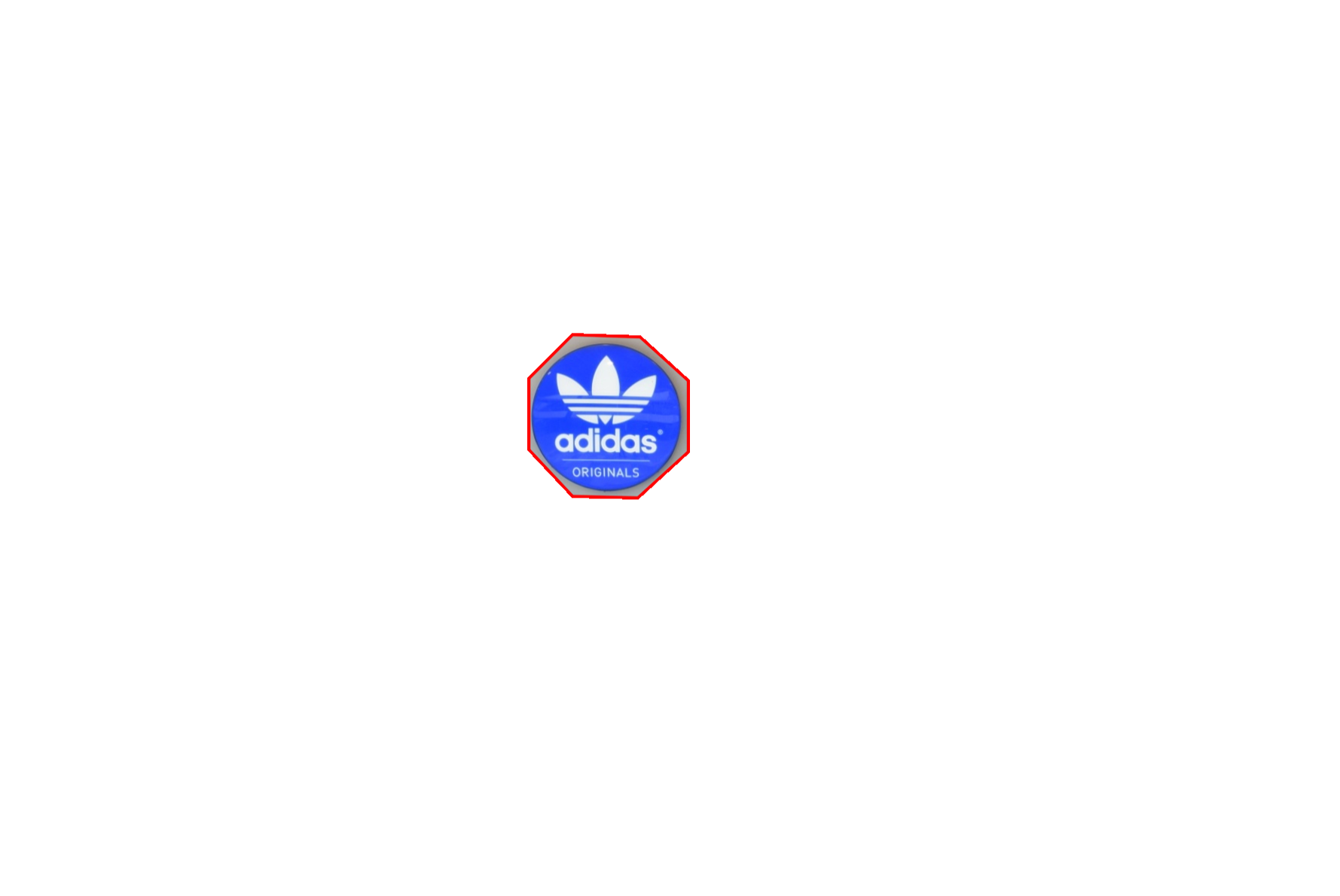} &
         \includegraphics[width=\scaleOOI\linewidth]{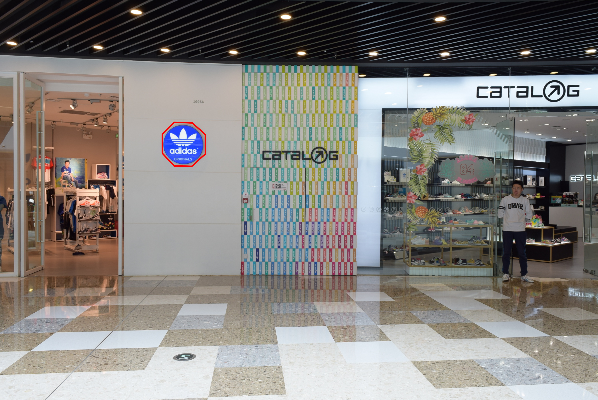}& \includegraphics[width=\scaleOOI\linewidth]{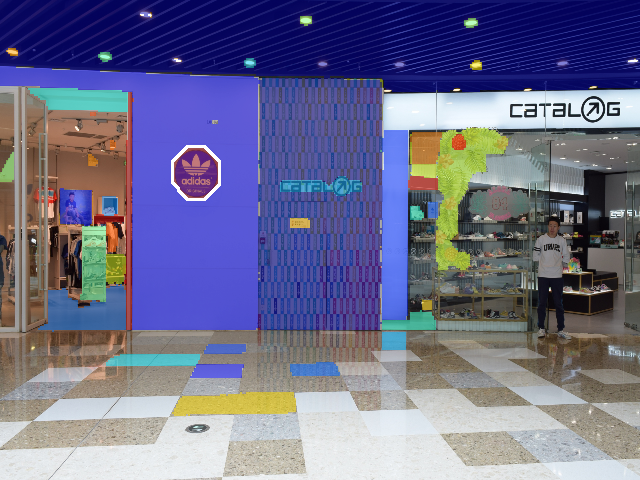} & \includegraphics[width=\scaleOOI\linewidth]{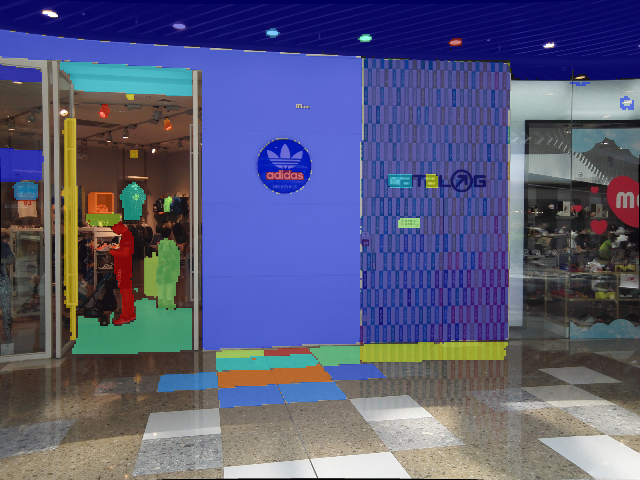}\\
         \includegraphics[width=\scaleOOI\linewidth,trim={5cm 5cm 5cm 5cm},clip]{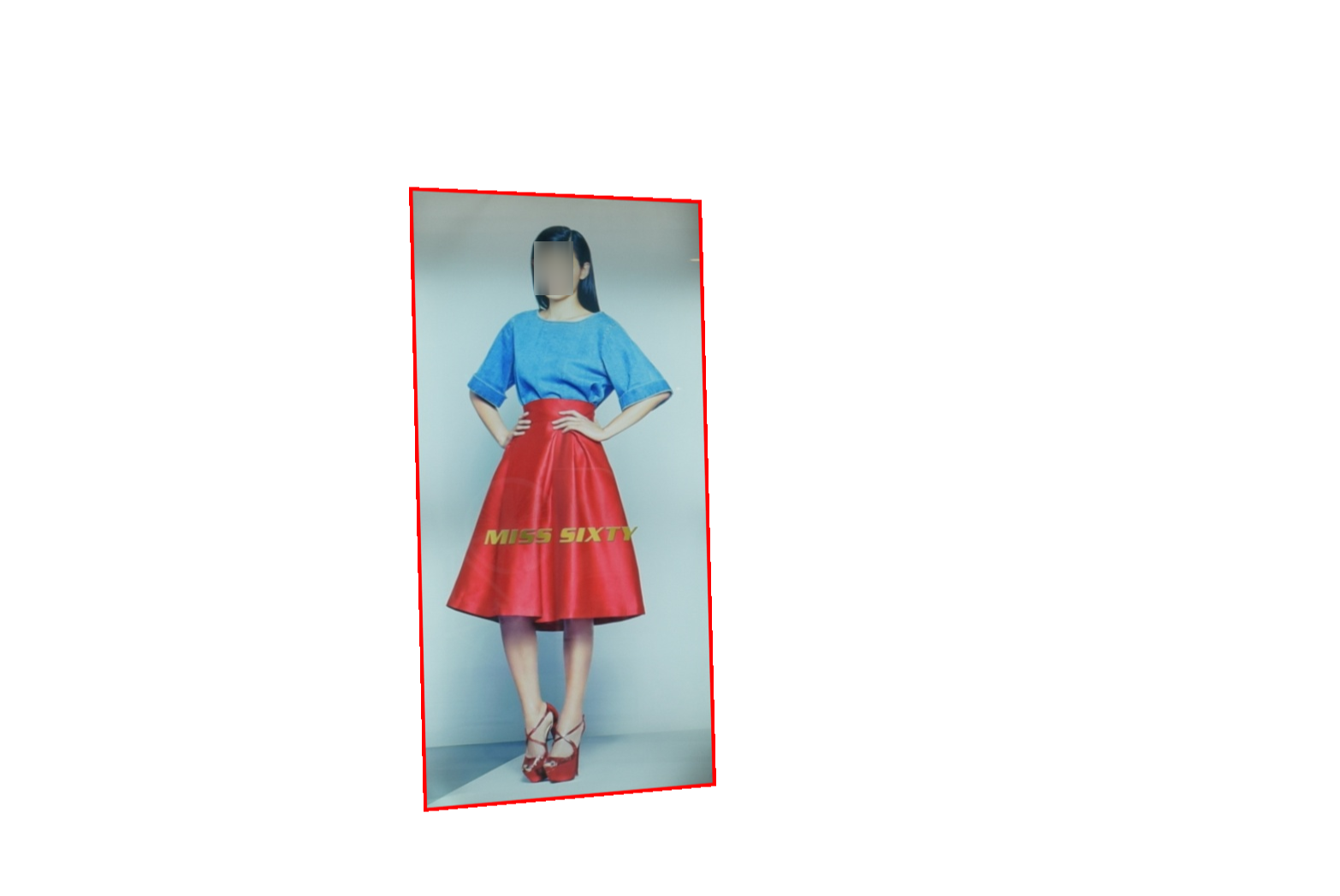} &
          \includegraphics[width=\scaleOOI\linewidth]{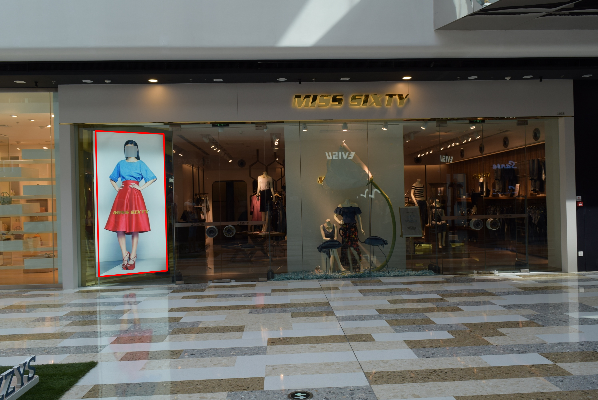}& \includegraphics[width=\scaleOOI\linewidth]{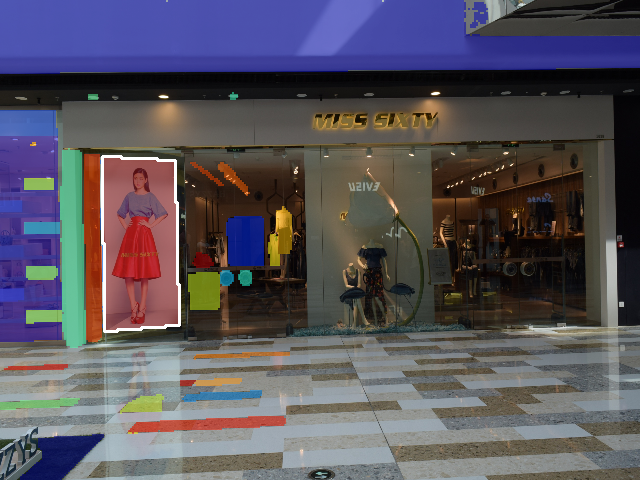} & \includegraphics[width=\scaleOOI\linewidth]{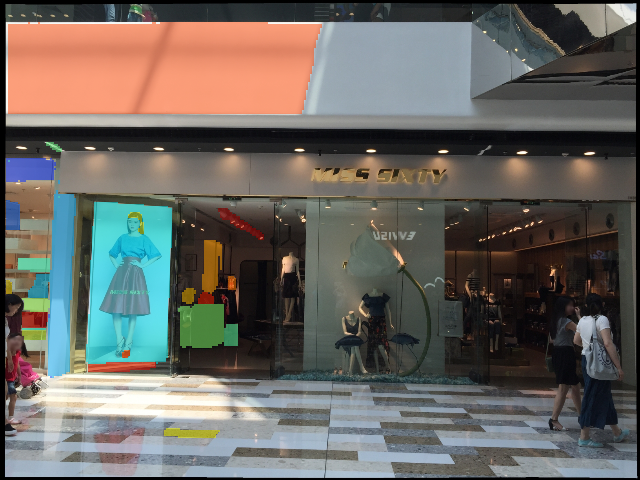}\\
           \includegraphics[width=\scaleOOI\linewidth,trim={0cm 5cm 12cm 5cm},clip]{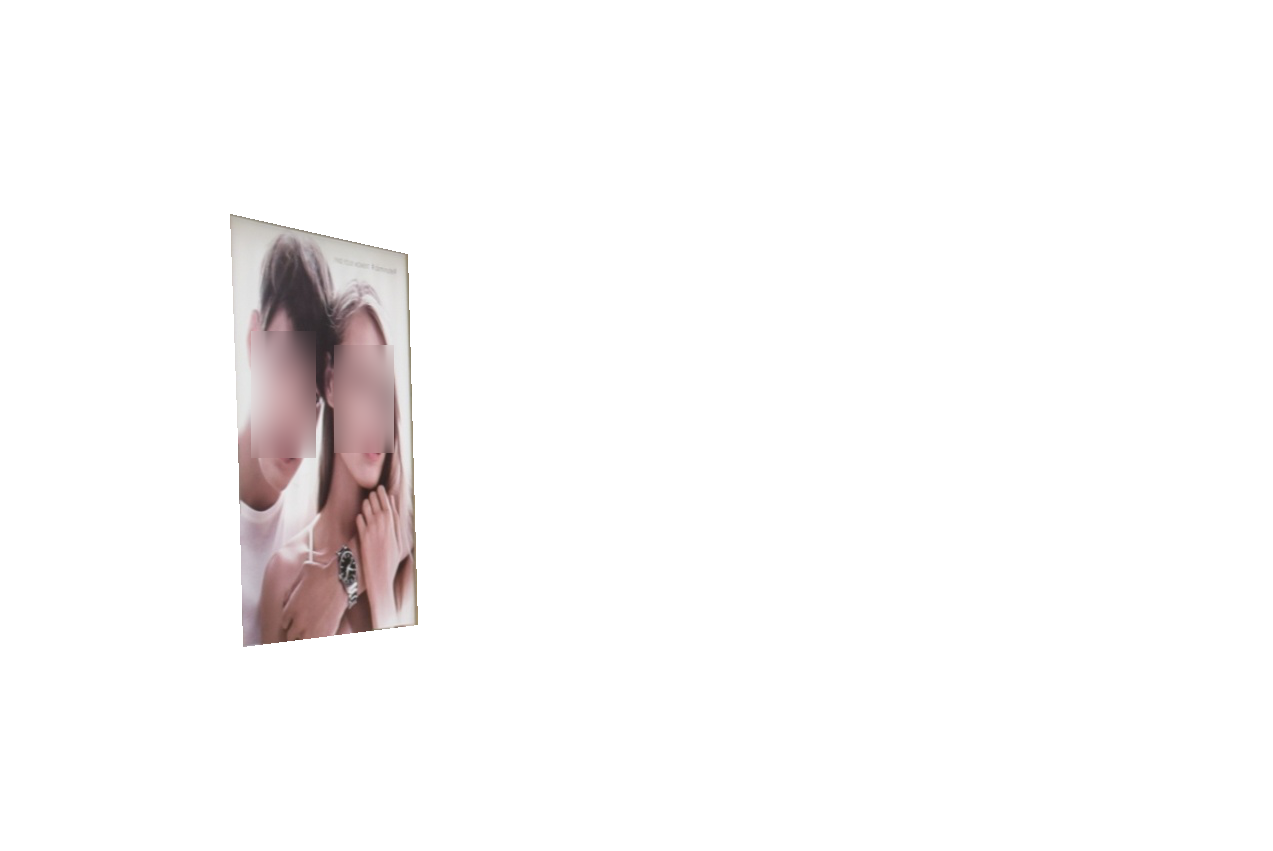} &
          \includegraphics[width=\scaleOOI\linewidth]{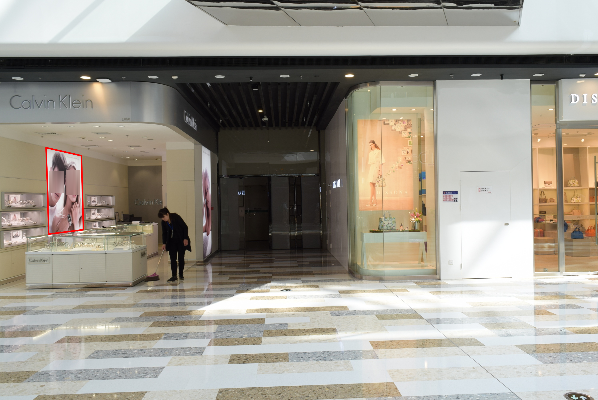}& \includegraphics[width=\scaleOOI\linewidth]{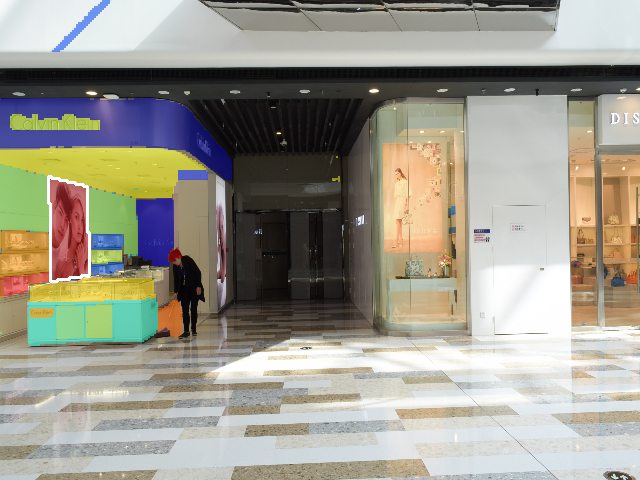} & \includegraphics[width=\scaleOOI\linewidth]{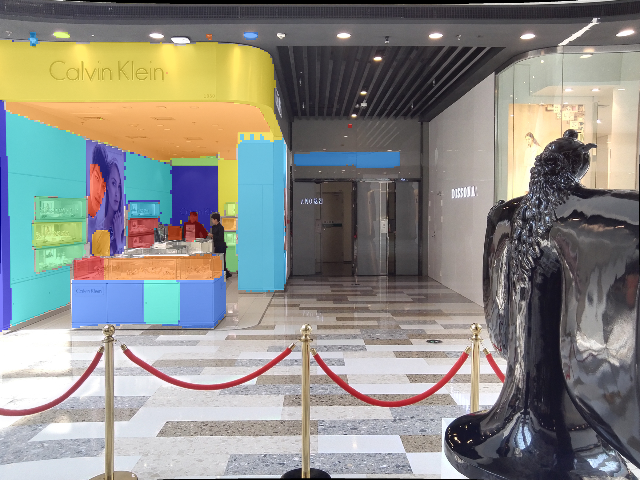}\\
           \includegraphics[width=\scaleOOI\linewidth,trim={12cm 10cm 12cm 12cm},clip]{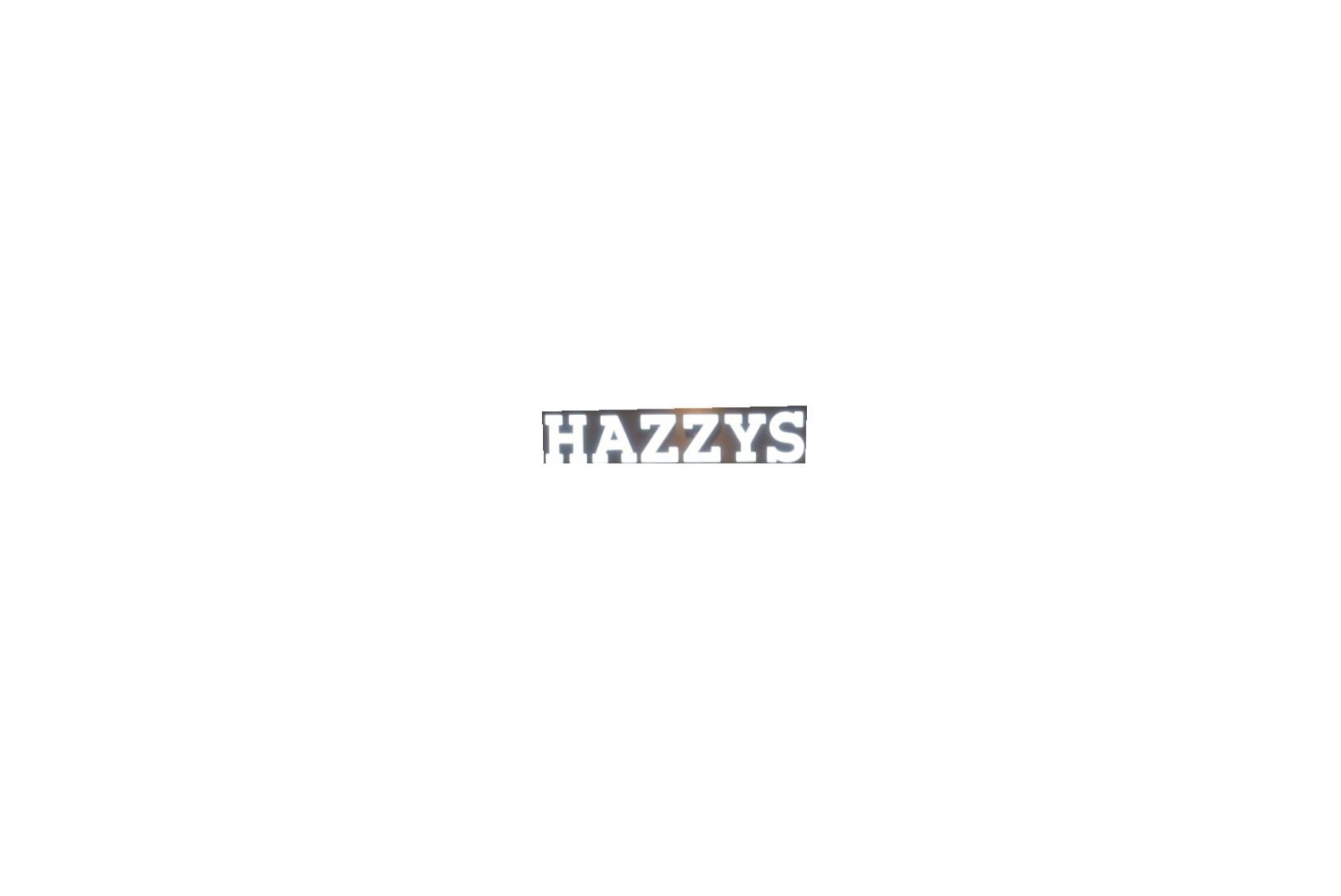} &
          \includegraphics[width=\scaleOOI\linewidth]{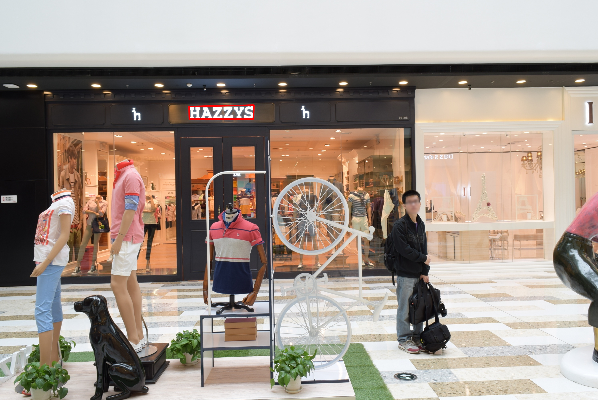}& \includegraphics[width=\scaleOOI\linewidth]{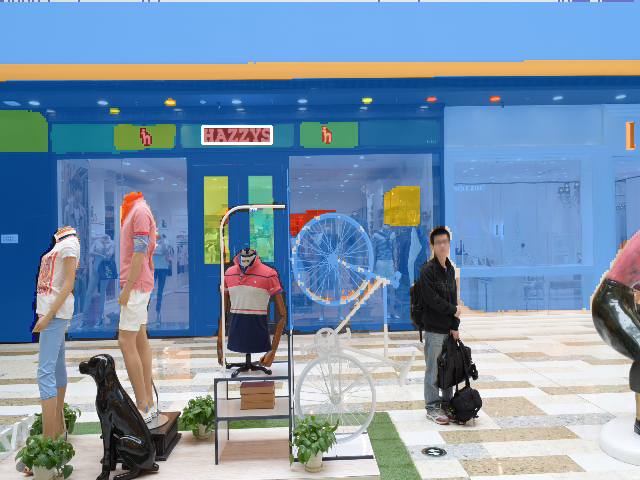} & \includegraphics[width=\scaleOOI\linewidth]{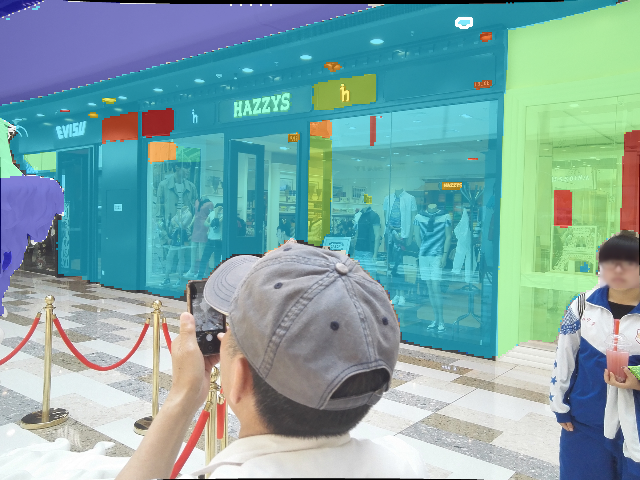}\\
    \end{tabular}%
    \caption{Qualitative results for OOI retrieval. Column 1 and 2 show OOI (marked in red bounding box). Column 3 shows the OOI as a segment along with its neighbors. Column 4 shows the matched segment.}
    \label{fig:ooi_qual}
\end{figure}

\clearpage

\section*{Acknowledgements}
This work was supported by the Centre for Augmented Reasoning (CAR) at the Australian Institute for Machine Learning (AIML), University of Adelaide, Australia. The authors acknowledge the computational support provided by the Indian Institute of Science (IISc), Bengaluru, India, and the
International Institute of Information Technology, Hyderabad (IIITH), India. The authors thank Ahmad Khaliq for technical support, Sarah Ibrahimi for sharing the InsideOut dataset, and Martin Humenberger for sharing the OOI annotations for the Baidu Mall dataset.

\bibliographystyle{splncs04}
\bibliography{main,egbib}
\end{document}